\documentclass[conference]{IEEEtran}

\IEEEoverridecommandlockouts

\usepackage{cite}
\usepackage{amsmath,amssymb,amsfonts}
\usepackage{algorithmic}
\usepackage{graphicx}
\usepackage{textcomp}
\usepackage{xcolor}
\usepackage{booktabs}
\usepackage{array}
\usepackage{adjustbox}
\usepackage{algorithm}
\usepackage{caption}
\usepackage{url}
\usepackage[hidelinks]{hyperref}

\def\BibTeX{{\rm B\kern-.05em{\sc i\kern-.025em b}\kern-.08em
    T\kern-.1667em\lower.7ex\hbox{E}\kern-.125emX}}

\begin{document}

\title{Detecting Temporally Localized Manipulations in Authentic Video Streams}

\author{
\IEEEauthorblockN{Okan Umur}
\IEEEauthorblockA{
Department of Software Engineering\\
Sakarya University\\
Sakarya, Türkiye\\
umuro2124@gmail.com
}
\and

\IEEEauthorblockN{Ali Emre G\"u\c{s}l\"u}
\IEEEauthorblockA{
Department of Software Engineering\\
Sakarya University\\
Sakarya, Türkiye\\
ali.guslu@ogr.sakarya.edu.tr
}
\and

\IEEEauthorblockN{{I}brahim Delibasoglu}
\IEEEauthorblockA{
Department of Software Engineering\\
Sakarya University\\
Sakarya, Türkiye\\
ibrahimdelibasoglu@sakarya.edu.tr
}
}

\maketitle

\begin{abstract}
The rapid advancement of video editing and generative artificial intelligence technologies has made realistic video manipulation increasingly accessible. Although existing datasets have significantly advanced research in deepfake detection, object removal, and video inpainting, they do not adequately model scenarios in which a short manipulated segment is inserted into an otherwise authentic video and the original video continues afterward. In this study, we review representative datasets from the literature, analyze their characteristics, and discuss their limitations with respect to temporally localized realistic manipulation detection. Based on this analysis, we motivate the need for a new dataset specifically designed for authentic videos containing short and highly realistic manipulated intervals. Finally, we evaluate two complementary approaches on our custom-curated test set to establish an initial benchmark for this challenging scenario. The first employs a linear probe on DINOv3 features, assessed under three thresholding strategies. The second leverages DINOv3 features with a consecutive frame similarity-based method to detect temporal manipulation boundaries. Together, these experiments provide an initial benchmark for partially manipulated video detection and highlight the need for content-adaptive thresholding mechanisms. The dataset, code, and supplementary materials are publicly available at
\href{https://github.com/OkanUmur/temporally-localized-video-manipulation-detection}
{GitHub repository}.
\end{abstract}


% ============================================================
\section{Introduction}
% ============================================================

The rapid advancement of digital media technologies has significantly increased the accessibility of video editing tools, enabling users to manipulate video content with high levels of realism. While these technologies provide important benefits for entertainment, film production, and content creation, they also raise serious concerns regarding digital media authenticity, misinformation, and forensic reliability. As a result, detecting manipulated video content has become an important research problem in multimedia forensics and computer vision.

Recent studies in video forgery detection have primarily focused on deepfake manipulations, particularly those involving facial identity replacement, facial reenactment, or synthetic speech generation. Large-scale datasets such as LAV-DF \cite{lavdf} and AV-Deepfake1M \cite{avdeepfake1m} have been introduced to support research on audio-visual deepfake detection and multimodal forgery localization. In addition, several datasets have been proposed for object-level video manipulations, including object removal and video inpainting scenarios. For example, the TVIL dataset \cite{Zhang_2023} focuses on temporal localization of video inpainting artifacts, while datasets such as DAVIS \cite{davis2016}, VideoSham \cite{videosham}, VPData/VPBench \cite{vpdata}, FSVD \cite{fsvd}, and CSVTED \cite{csvted} provide manipulated videos generated through object removal, insertion, or general scene editing.

Although these datasets have significantly contributed to the development of video forgery detection methods, they often focus on fully manipulated videos, face-centric deepfakes, object removal, or audio manipulation scenarios. In many real-world situations, however, video manipulation may occur in a more subtle manner. A short manipulated segment may be inserted into an otherwise authentic video while preserving the temporal continuity of the original footage. In such cases, the manipulated content must visually blend with the surrounding authentic frames, making detection substantially more challenging.

Despite the growing body of research on video forgery detection, datasets that explicitly model temporally localized manipulations embedded within authentic videos remain limited. Most existing datasets do not provide scenarios where a video begins with authentic content, contains a short manipulated segment, and then continues with the original authentic sequence. This type of manipulation represents a realistic tampering scenario that can occur in surveillance footage, social media videos, edited reports, or event recordings.

To address this limitation, this study investigates the problem of detecting temporally localized manipulated segments inserted into authentic video streams. We first analyze widely used datasets in the literature and identify their limitations with respect to this specific manipulation pattern. Based on this analysis, we motivate the need for a new dataset and construct a custom-curated test set to evaluate a baseline detection framework under this challenging scenario.

The main contributions of this work can be summarized as follows:
\begin{itemize}
    \item We provide a comparative review of existing video forgery datasets 
    relevant to temporal and object-level manipulation analysis.
    \item We identify the lack of datasets that explicitly model short manipulated 
    segments inserted into otherwise authentic video streams.
    \item We construct a custom-curated, partially manipulated video test set to 
    represent this realistic tampering scenario.
    \item We propose and evaluate a supervised baseline using a linear probe on 
    self-supervised DINOv3~\cite{simeoni2025dinov3} features under three thresholding strategies, 
    establishing an initial benchmark for future work.
    \item We propose and evaluate an unsupervised temporal anomaly detection 
    framework based on consecutive frame cosine similarity of DINOv3 features, 
    achieving 83\% precision and 95\% video-level accuracy on authentic streams 
    without any training.
\end{itemize}

% ============================================================
\section{Related Datasets}
% ============================================================

Several datasets have been proposed to support research on video forgery detection, deepfake analysis, temporal localization, and object-level manipulation detection. In this section, we summarize the most relevant datasets considered in our study and discuss their limitations with respect to our target scenario.

\subsection{LAV-DF}
\begin{figure}[htbp]
    \centering
    \includegraphics[width=0.48\columnwidth]{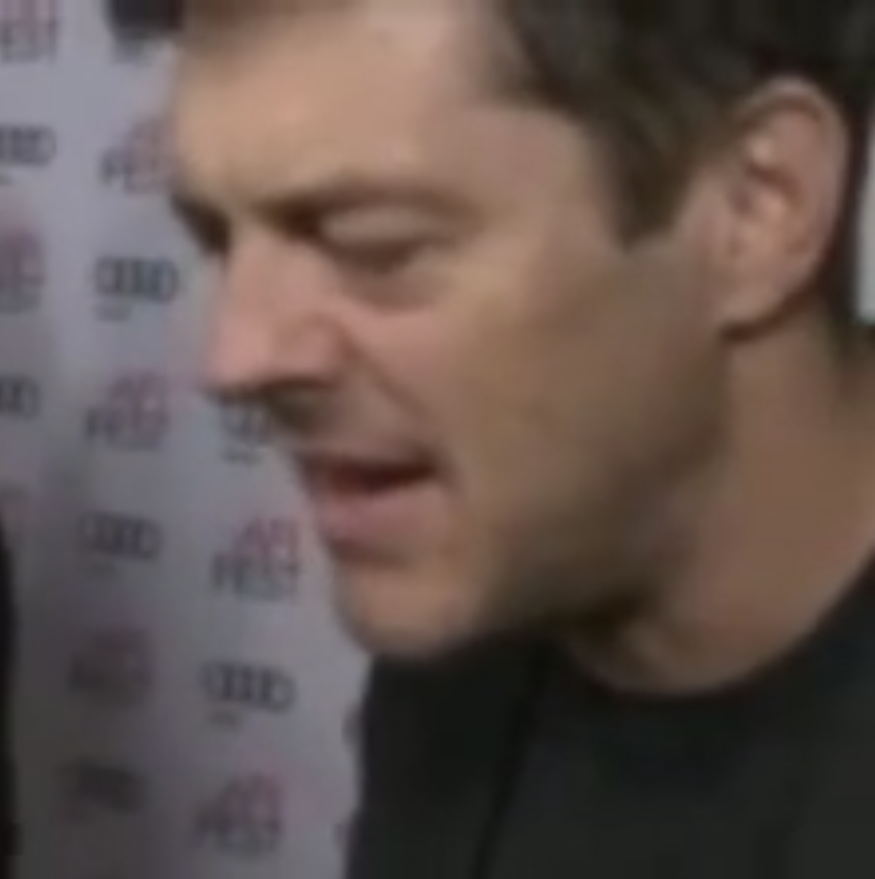}
    \includegraphics[width=0.48\columnwidth]{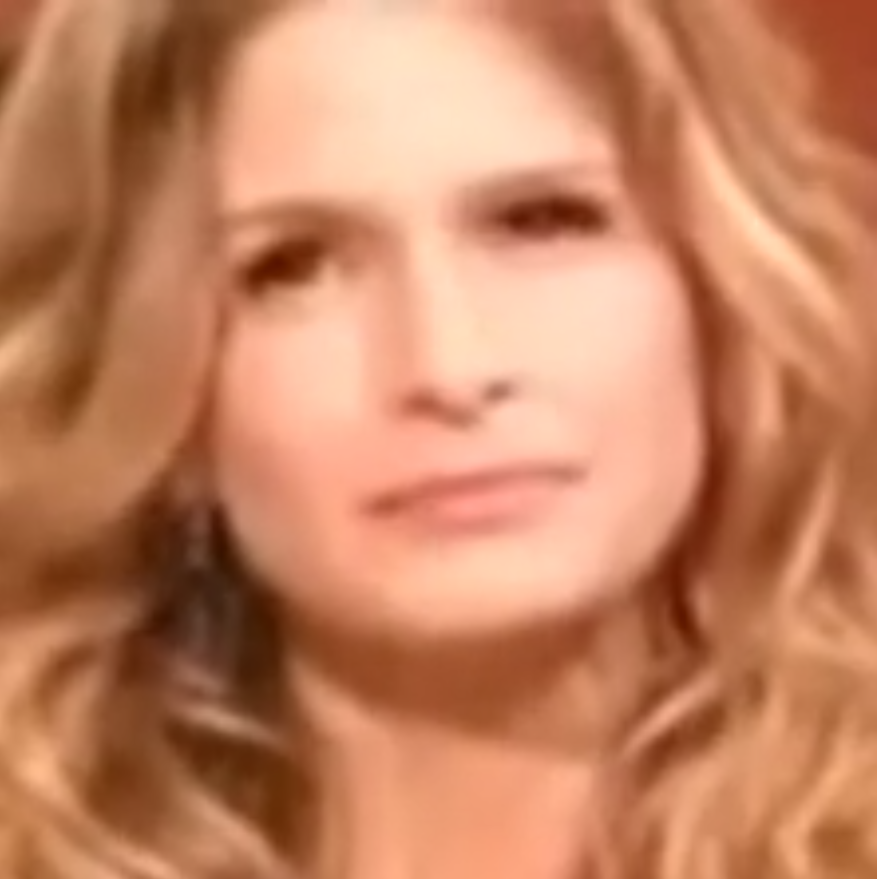}
    \caption{Example frames illustrating the LAV-DF manipulation scenario. The dataset mainly focuses on face-centric audio-visual manipulations, which differs from our target scenario of short manipulated intervals inserted into otherwise authentic videos.}
    \label{fig:lavdf_examples}
\end{figure}

LAV-DF was introduced by Cai et al. \cite{lavdf} for audio-visual deepfake detection and temporal forgery localization. The dataset contains a large number of real and manipulated videos with visual-only, audio-only, and joint audio-visual manipulations. Although LAV-DF provides temporal annotations, it mainly focuses on face-centric manipulations and therefore does not directly represent general object-level manipulations embedded within authentic videos. As illustrated in Figure~\ref{fig:lavdf_examples}, the manipulation scenario in LAV-DF is primarily centered on facial and audio-visual deepfake content rather than general scene-level inserted manipulations.

\subsection{AV-Deepfake1M}
\begin{figure}[htbp]
    \centering
    \includegraphics[width=0.24\columnwidth]{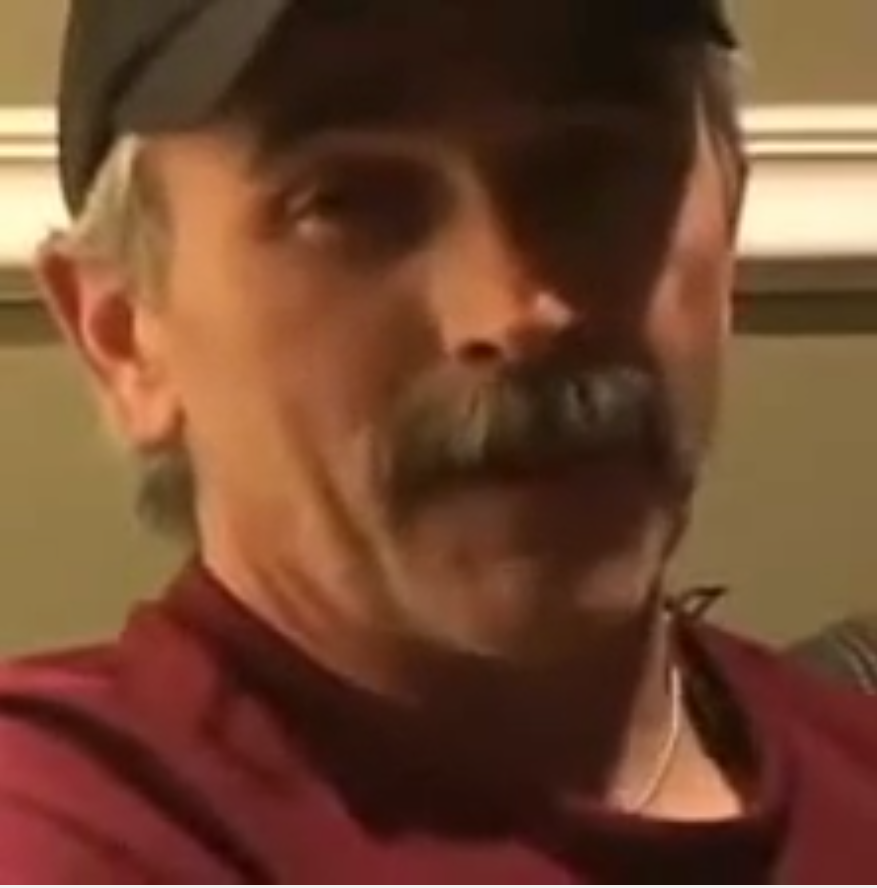}
    \includegraphics[width=0.24\columnwidth]{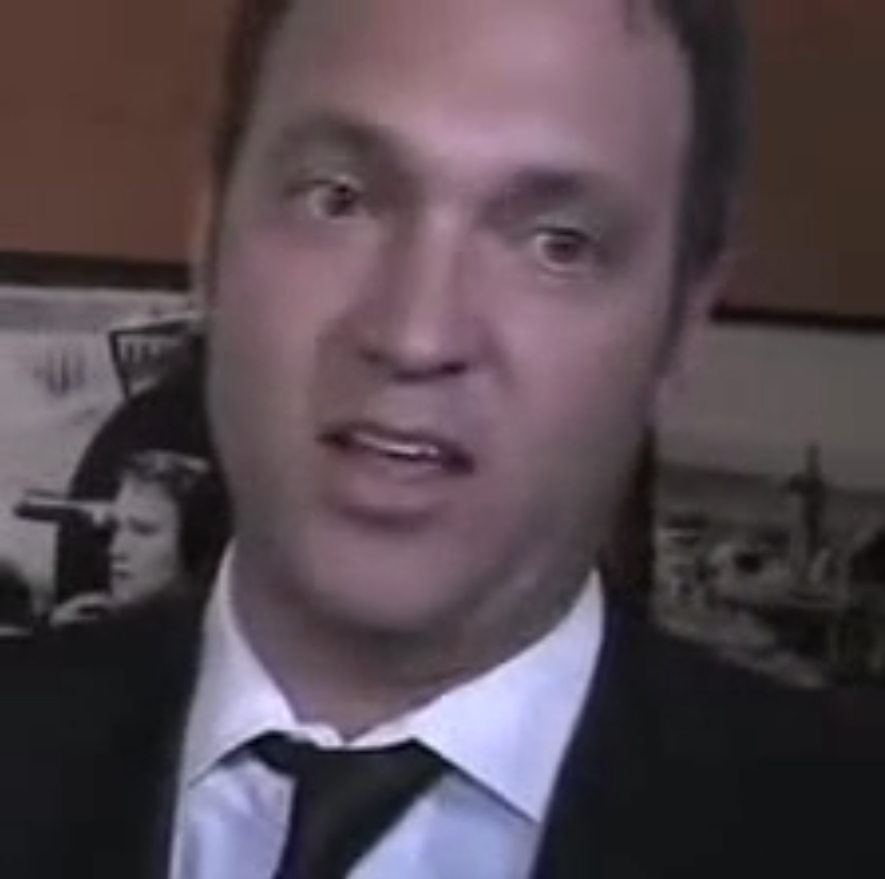}
    \includegraphics[width=0.24\columnwidth]{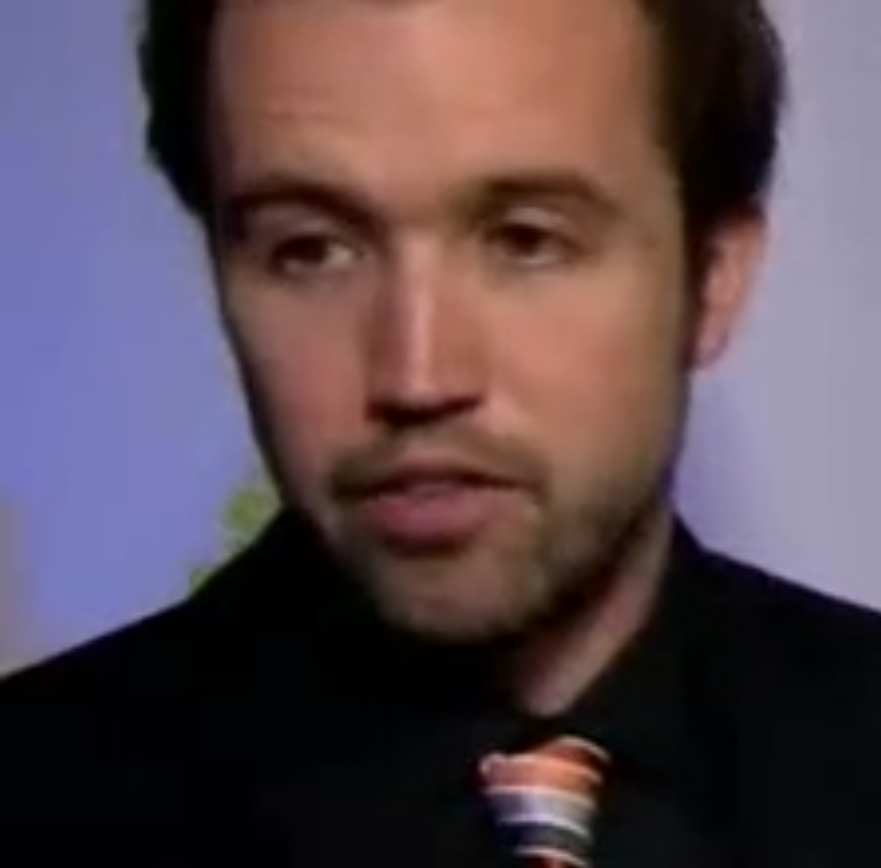}
    \includegraphics[width=0.24\columnwidth]{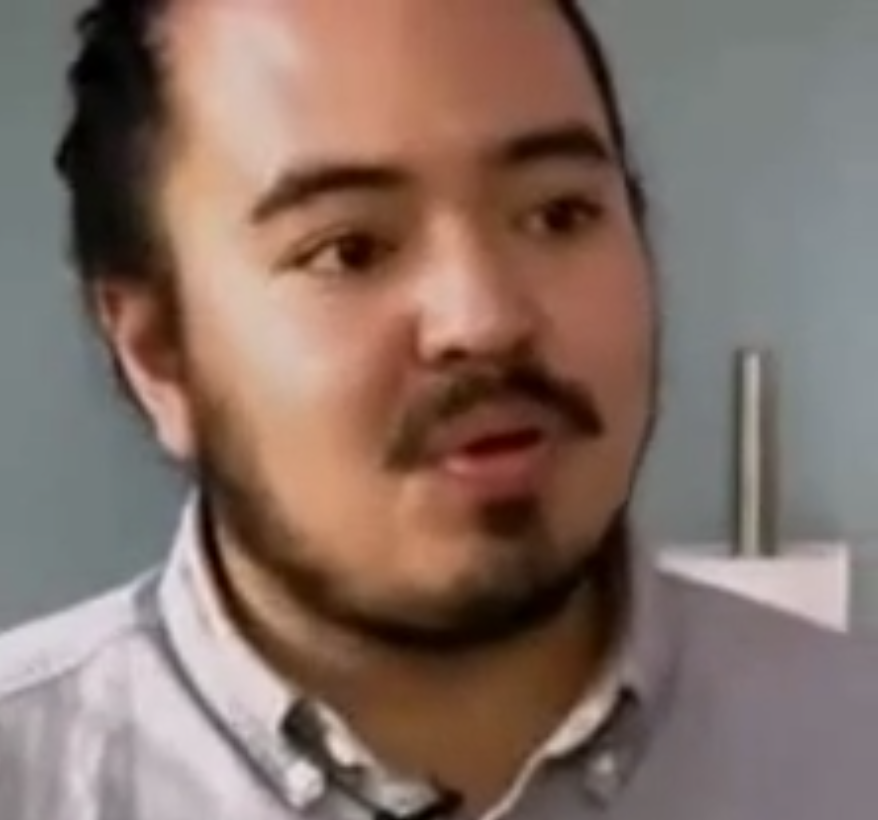}
    \caption{Example frames illustrating the AV-Deepfake1M manipulation scenario. The dataset primarily focuses on large-scale identity-based audio-visual deepfake generation rather than short manipulated intervals inserted into otherwise authentic video streams.}
    \label{fig:avdeepfake_examples}
\end{figure}

AV-Deepfake1M \cite{avdeepfake1m} is a large-scale multimodal deepfake dataset designed for audio-visual forgery detection. Its scale is beneficial for training deep learning models; however, it primarily focuses on facial and speech manipulations rather than short manipulated segments inserted into otherwise authentic video streams. As illustrated in Figure~\ref{fig:avdeepfake_examples}, AV-Deepfake1M mainly represents identity-based audio-visual deepfake manipulations, which differ from our target scenario of inserting a short manipulated segment into an otherwise authentic video.

\subsection{TVIL}
\begin{figure}[htbp]
    \centering
    \includegraphics[width=0.48\columnwidth]{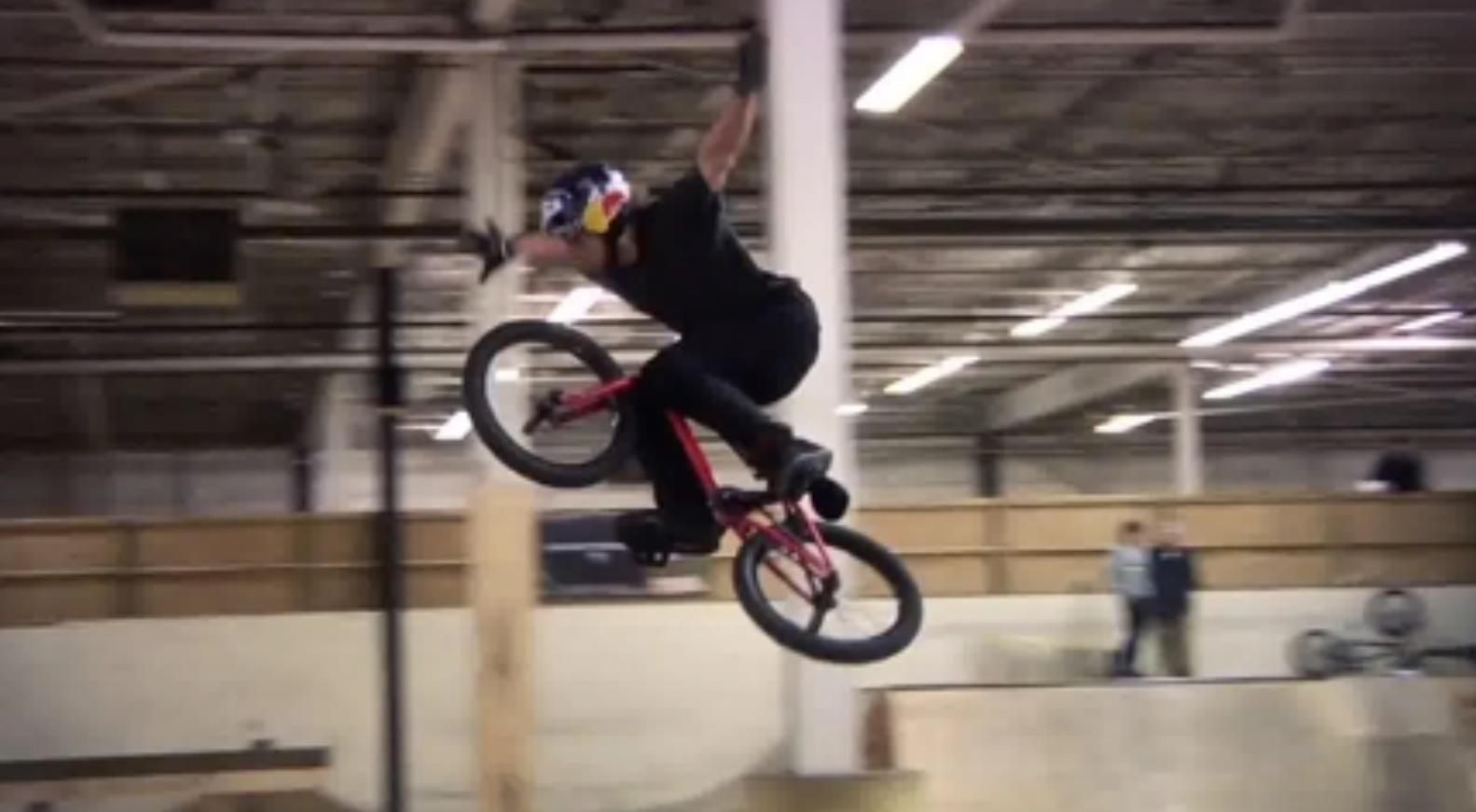}
    \includegraphics[width=0.48\columnwidth]{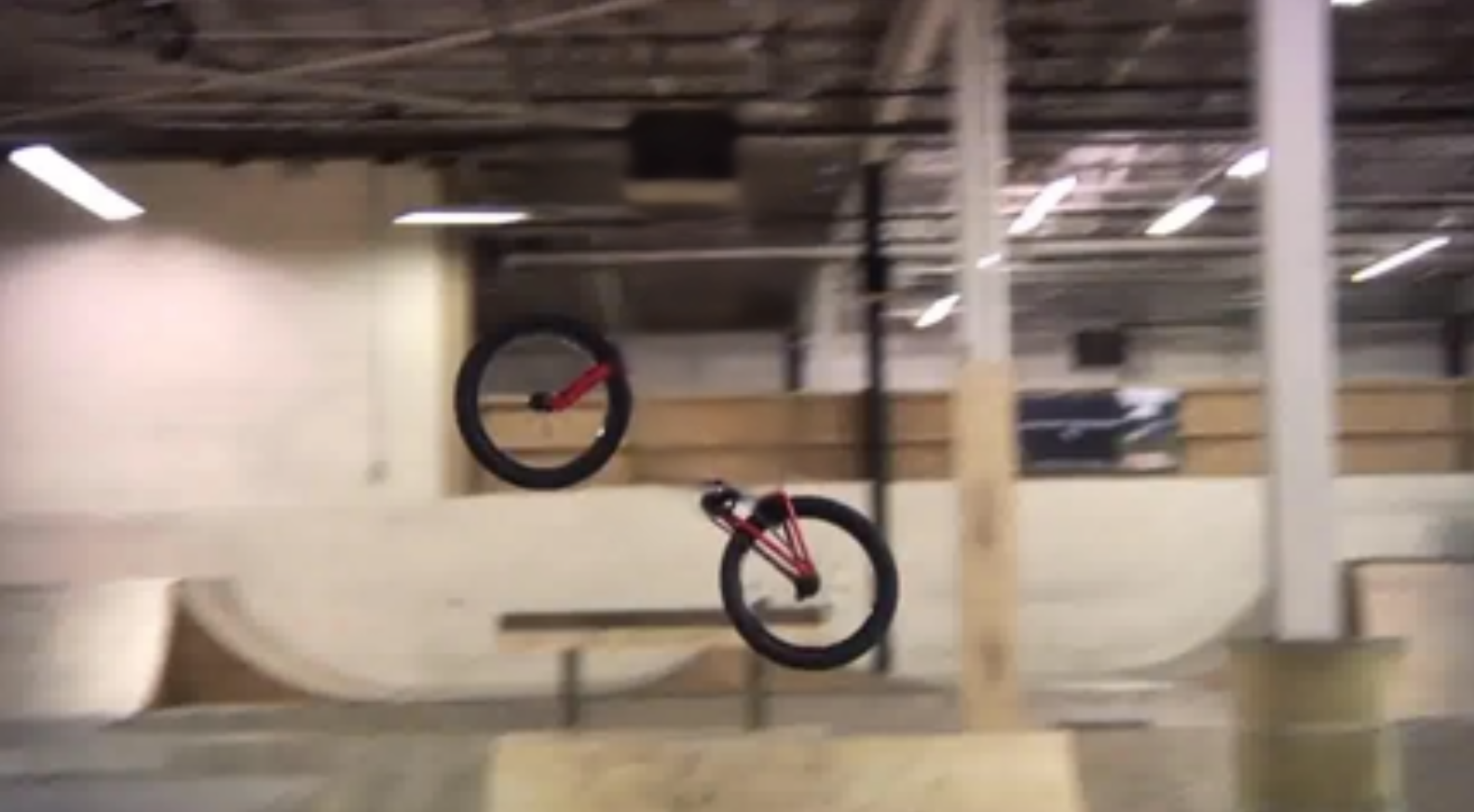}

    \vspace{2mm}

    \includegraphics[width=0.48\columnwidth]{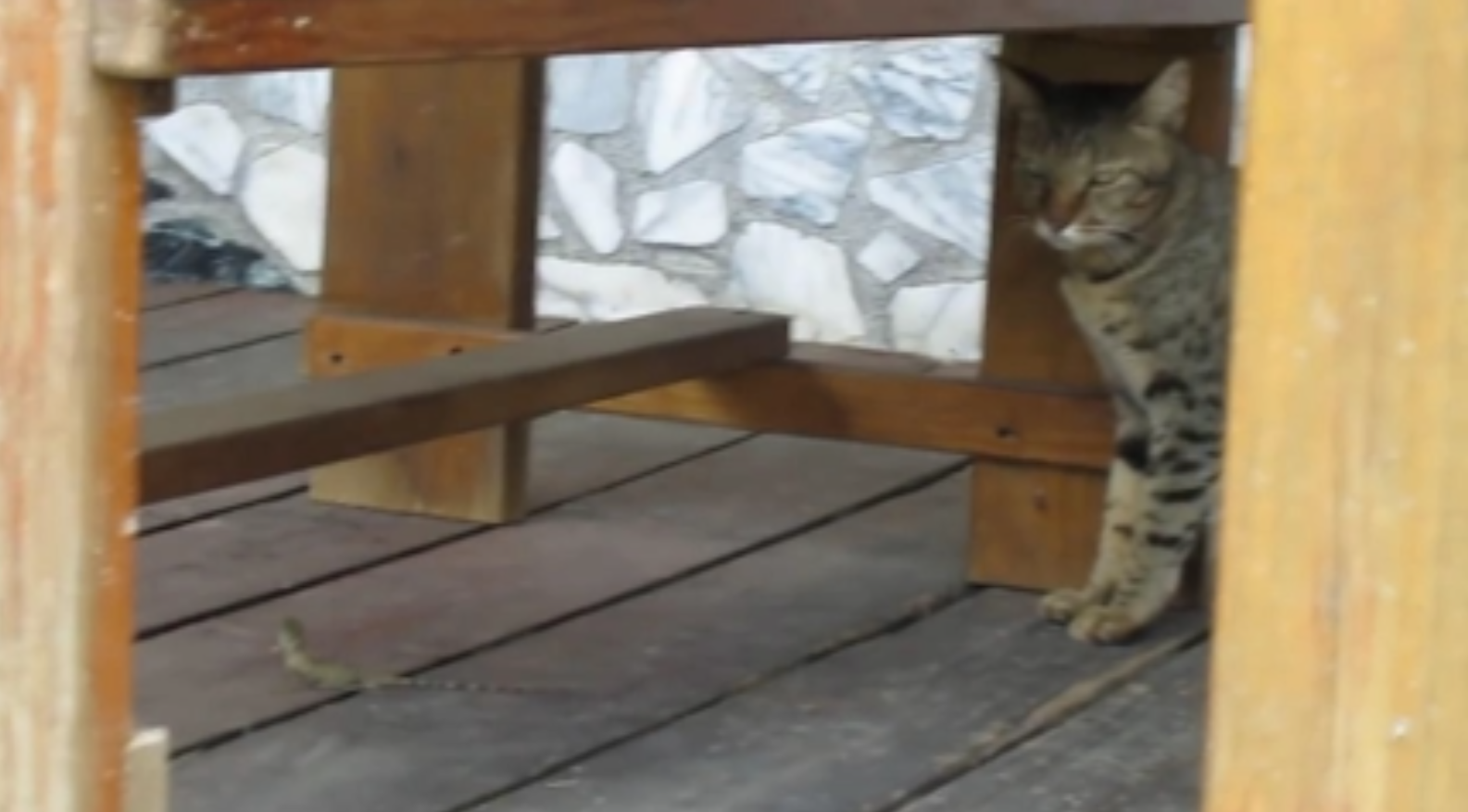}
    \includegraphics[width=0.48\columnwidth]{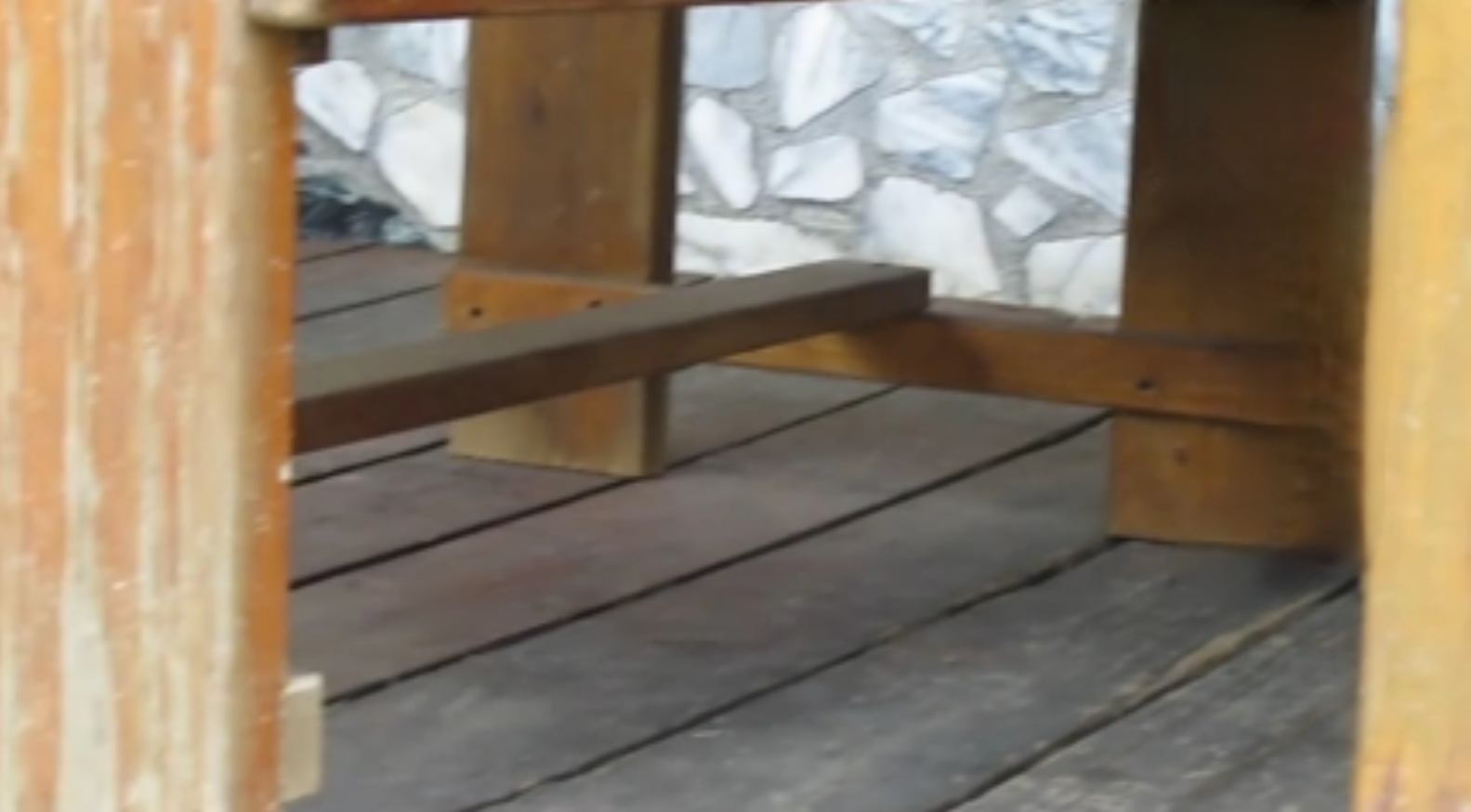}
    \caption{Example frames illustrating the TVIL manipulation scenario. The dataset focuses on temporal video inpainting localization and object removal artifacts rather than short manipulated segments inserted into otherwise authentic video streams.}
    \label{fig:tvil_examples}
\end{figure}

TVIL \cite{Zhang_2023} was proposed for temporal video inpainting localization and focuses on detecting manipulated object regions in general scenes. Compared with face-centric deepfake datasets, TVIL is more relevant to general scene manipulation. However, its primary emphasis is object removal via inpainting rather than realistic short inserted manipulations within a larger authentic video. As illustrated in Figure~\ref{fig:tvil_examples}, TVIL mainly captures object removal and inpainting-based manipulation patterns, which differ from our target scenario involving a short manipulated interval embedded within an otherwise authentic video.

\subsection{DAVIS-VI}
\begin{figure}[htbp]
    \centering
    \includegraphics[width=0.25\columnwidth,height=2.0cm]{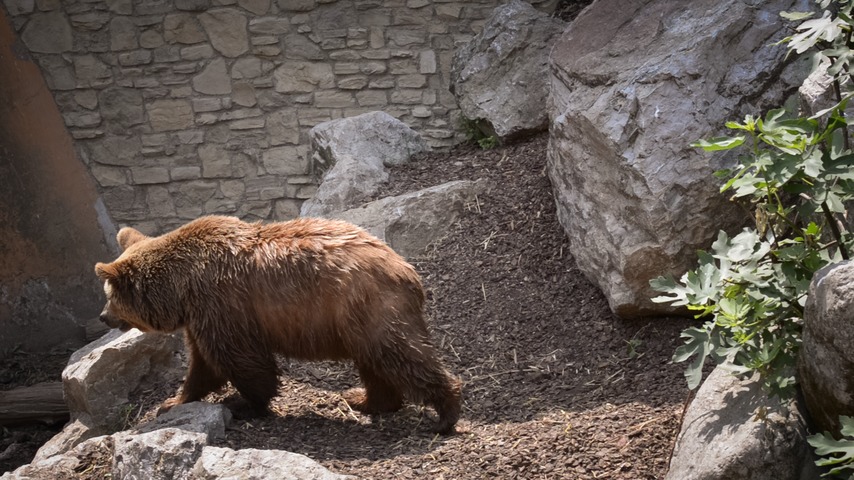}
    \hfill
    \includegraphics[width=0.25\columnwidth,height=2.0cm]{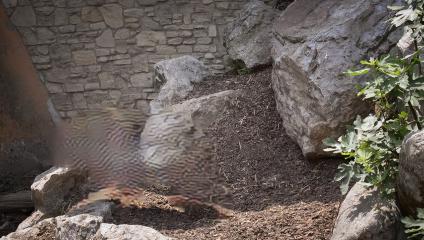}
    \hfill
    \includegraphics[width=0.25\columnwidth,height=2.0cm]{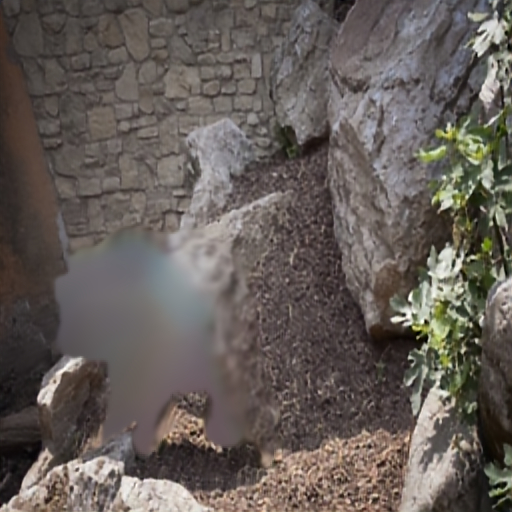}

    \vspace{2mm}

    \includegraphics[width=0.25\columnwidth,height=2.0cm]{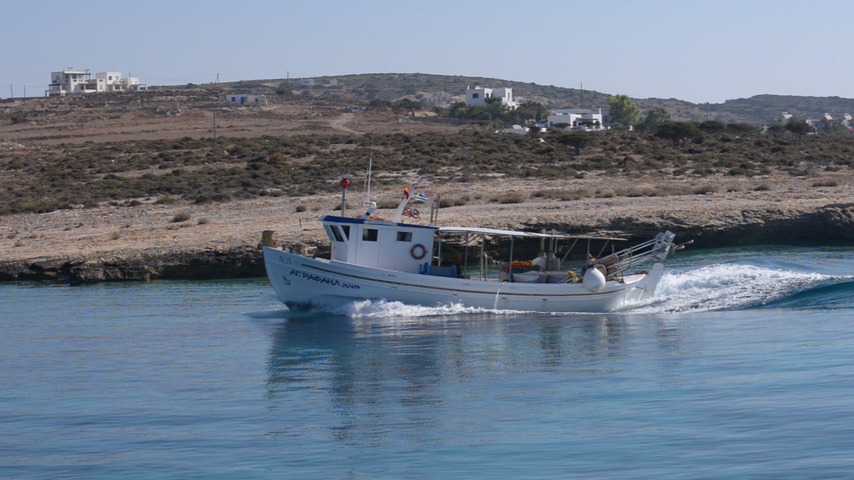}
    \hfill
    \includegraphics[width=0.25\columnwidth,height=2.0cm]{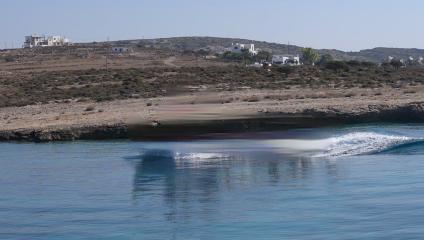}
    \hfill
    \includegraphics[width=0.25\columnwidth,height=2.0cm]{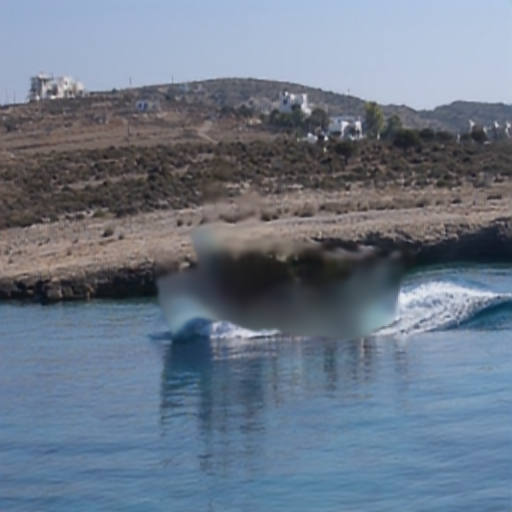}
    \caption{Example frames illustrating DAVIS-based inpainting scenarios. DAVIS-derived datasets are commonly used as a basis for object removal and inpainting generation, but they are not specifically designed to model short manipulated intervals inserted into otherwise authentic video streams.}
    \label{fig:davis_examples}
\end{figure}

DAVIS-based inpainting datasets are derived from the DAVIS benchmark introduced by Perazzi et al. \cite{davis2016}. In these datasets, moving objects are removed and missing regions are filled using video inpainting methods. These datasets are useful for studying inpainting artifacts, but they are not specifically designed for temporally localized inserted manipulations. As illustrated in Figure~\ref{fig:davis_examples}, DAVIS-based manipulation examples mainly reflect object removal and inpainting generation settings rather than the insertion of a short manipulated segment into an otherwise authentic video.

\subsection{VideoSham}
\begin{figure}[htbp]
    \centering
    \includegraphics[width=0.48\columnwidth,height=2.8cm]{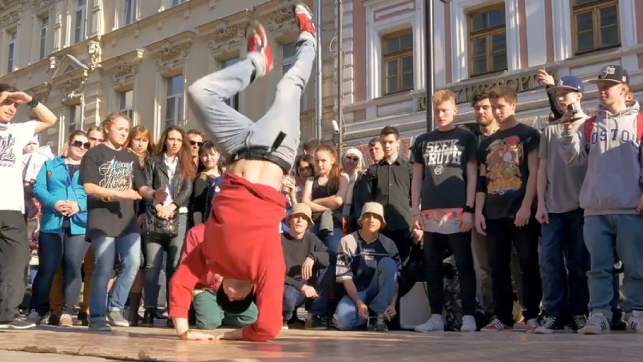}
    \hfill
    \includegraphics[width=0.48\columnwidth,height=2.8cm]{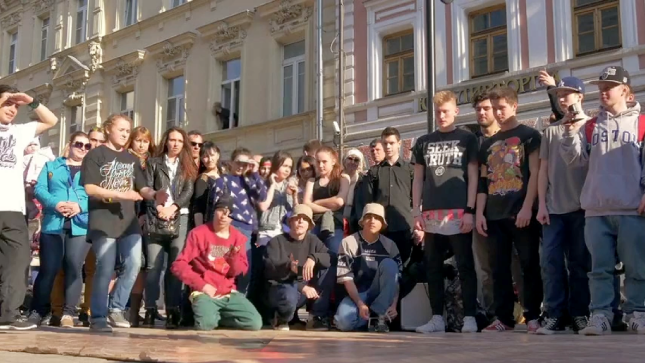}

    \vspace{2mm}

    \includegraphics[width=0.48\columnwidth,height=2.8cm]{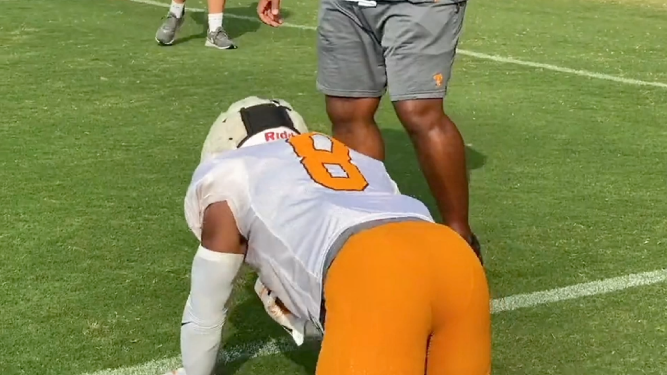}
    \hfill
    \includegraphics[width=0.48\columnwidth,height=2.8cm]{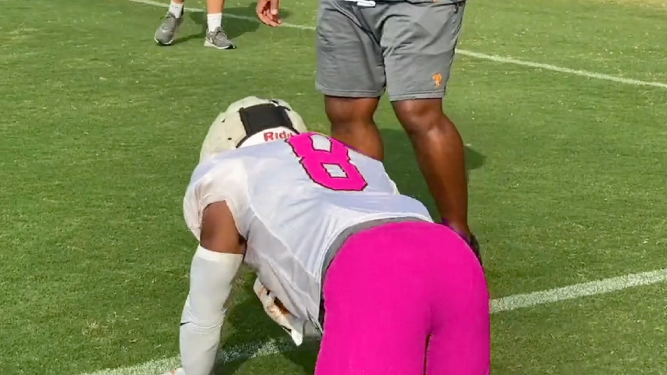}

    \vspace{2mm}

    \includegraphics[width=0.48\columnwidth,height=2.8cm]{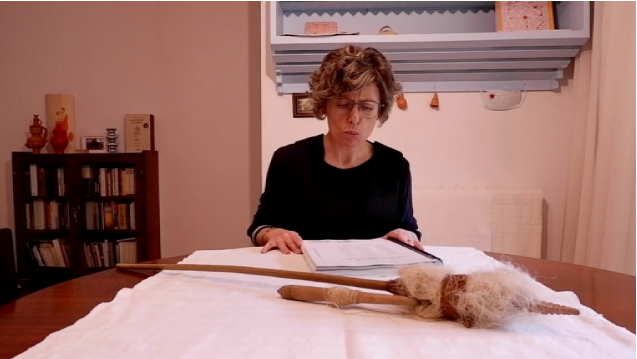}
    \hfill
    \includegraphics[width=0.48\columnwidth,height=2.8cm]{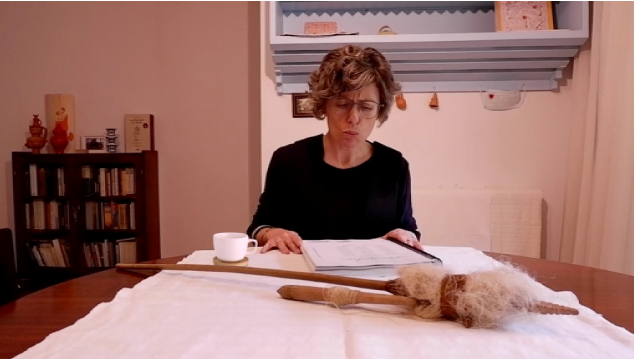}
    \caption{Example frames illustrating the VideoSham manipulation scenario. The dataset includes general video manipulations such as object insertion and removal, but it does not explicitly model short manipulated intervals inserted into otherwise authentic video streams.}
    \label{fig:videosham_examples}
\end{figure}

VideoSham \cite{videosham} focuses on general video manipulations beyond face deepfakes, including object insertion, object removal, and AI-based content editing. It is more relevant to real-world tampering than many face-only datasets. However, it does not explicitly model the scenario in which a short manipulated segment is inserted into an otherwise authentic video and the original footage resumes afterward. As illustrated in Figure~\ref{fig:videosham_examples}, VideoSham includes general scene-level manipulations, yet it does not explicitly represent the scenario where a short manipulated interval is inserted into an otherwise authentic video and followed by the continuation of the original footage.

\subsection{VPData and VPBench}
VPData/VPBench \cite{vpdata} provide large-scale video inpainting data with dense annotations and segmentation masks. These datasets are useful for large-scale training in object removal and background reconstruction tasks. Nevertheless, they mainly target inpainting and do not directly represent short temporally localized manipulation insertions.

\subsection{FSVD}
FSVD \cite{fsvd} was created for real-world forged smartphone videos and includes object insertion and deletion operations under practical capture conditions such as camera shake, compression artifacts, and varying illumination. It is valuable for realistic evaluation, yet it does not explicitly focus on short inserted manipulated intervals within otherwise authentic video streams.

\subsection{CSVTED}
\begin{figure}[htbp]
    \centering
    \includegraphics[width=0.31\columnwidth,height=2.6cm]{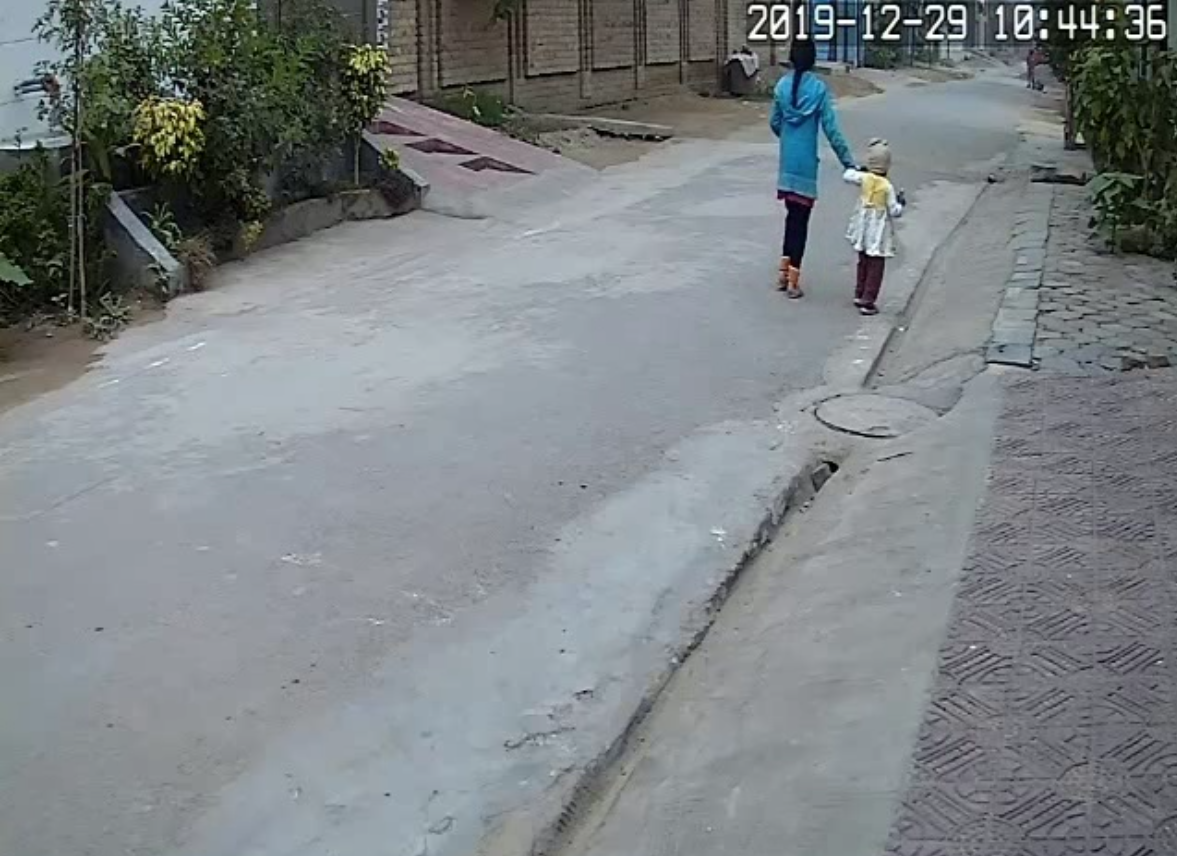}
    \hfill
    \includegraphics[width=0.31\columnwidth,height=2.6cm]{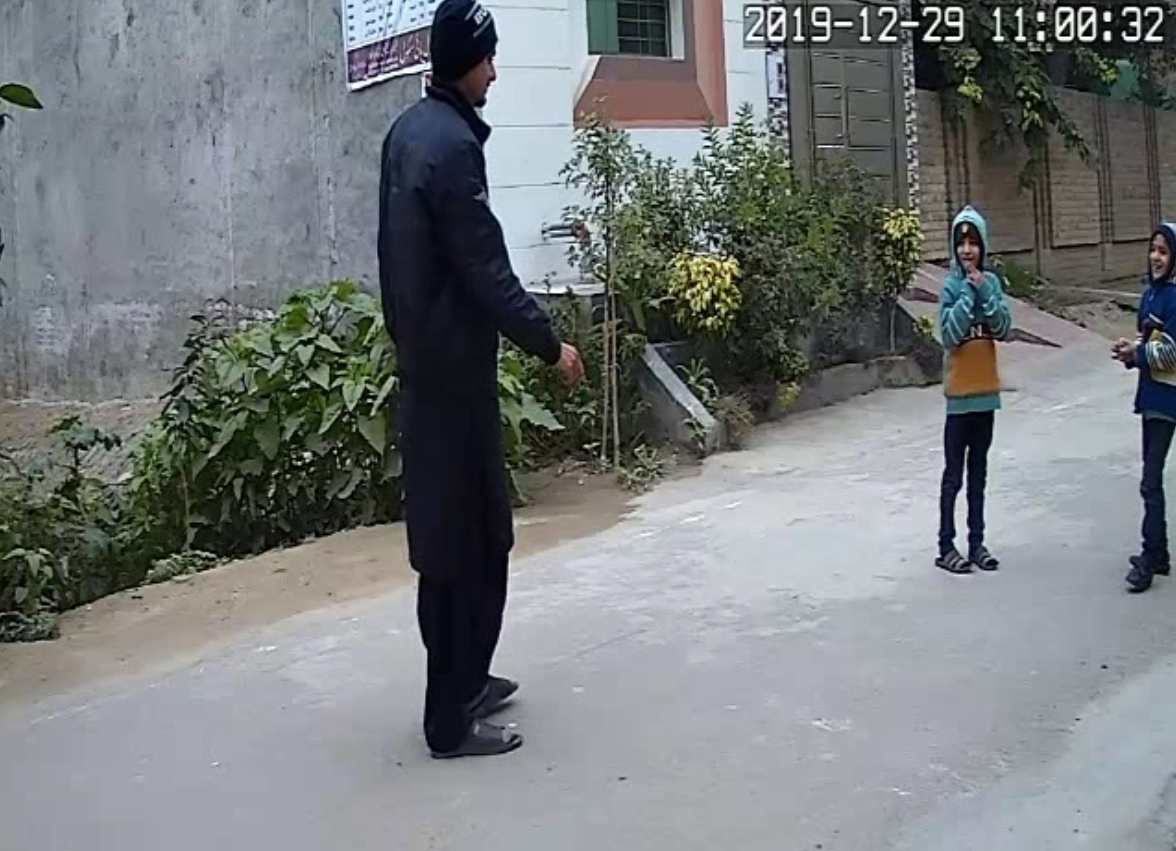}
    \hfill
    \includegraphics[width=0.31\columnwidth,height=2.6cm]{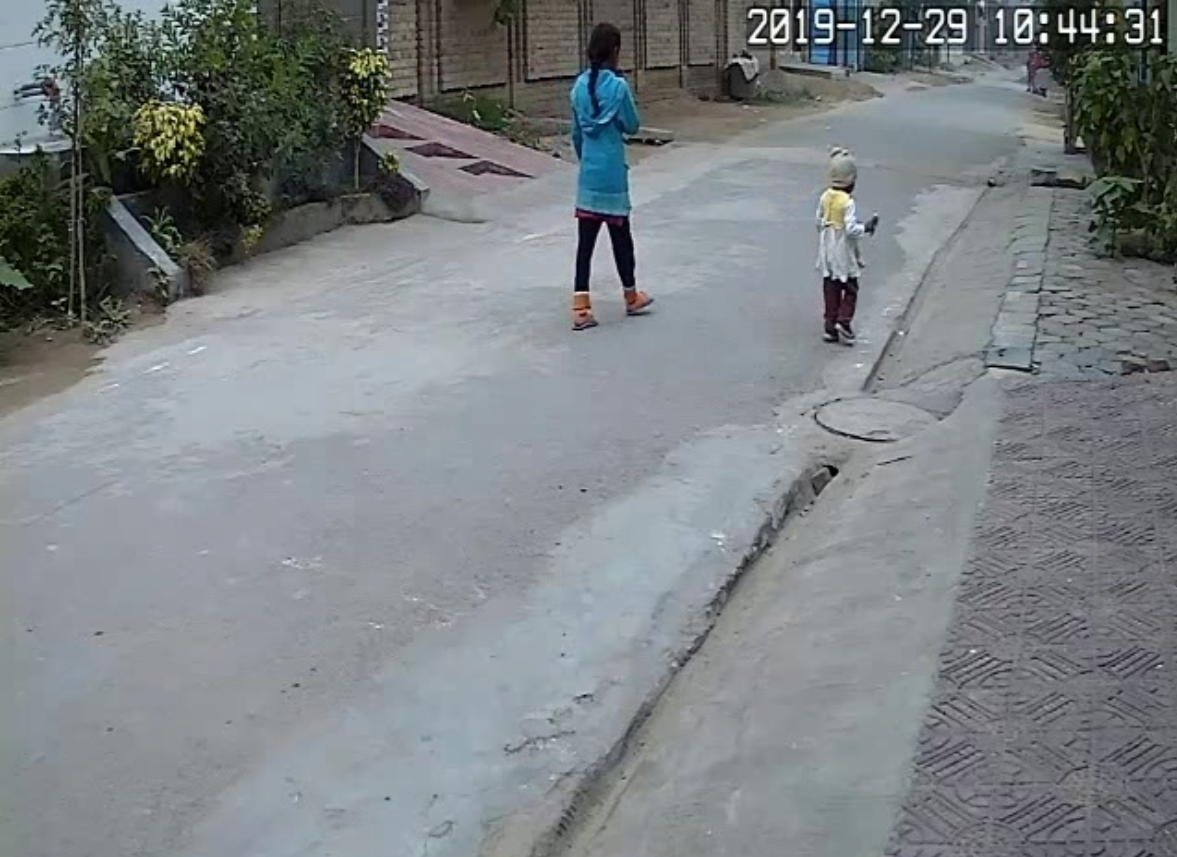}

    \vspace{2mm}

    \includegraphics[width=0.31\columnwidth,height=2.6cm]{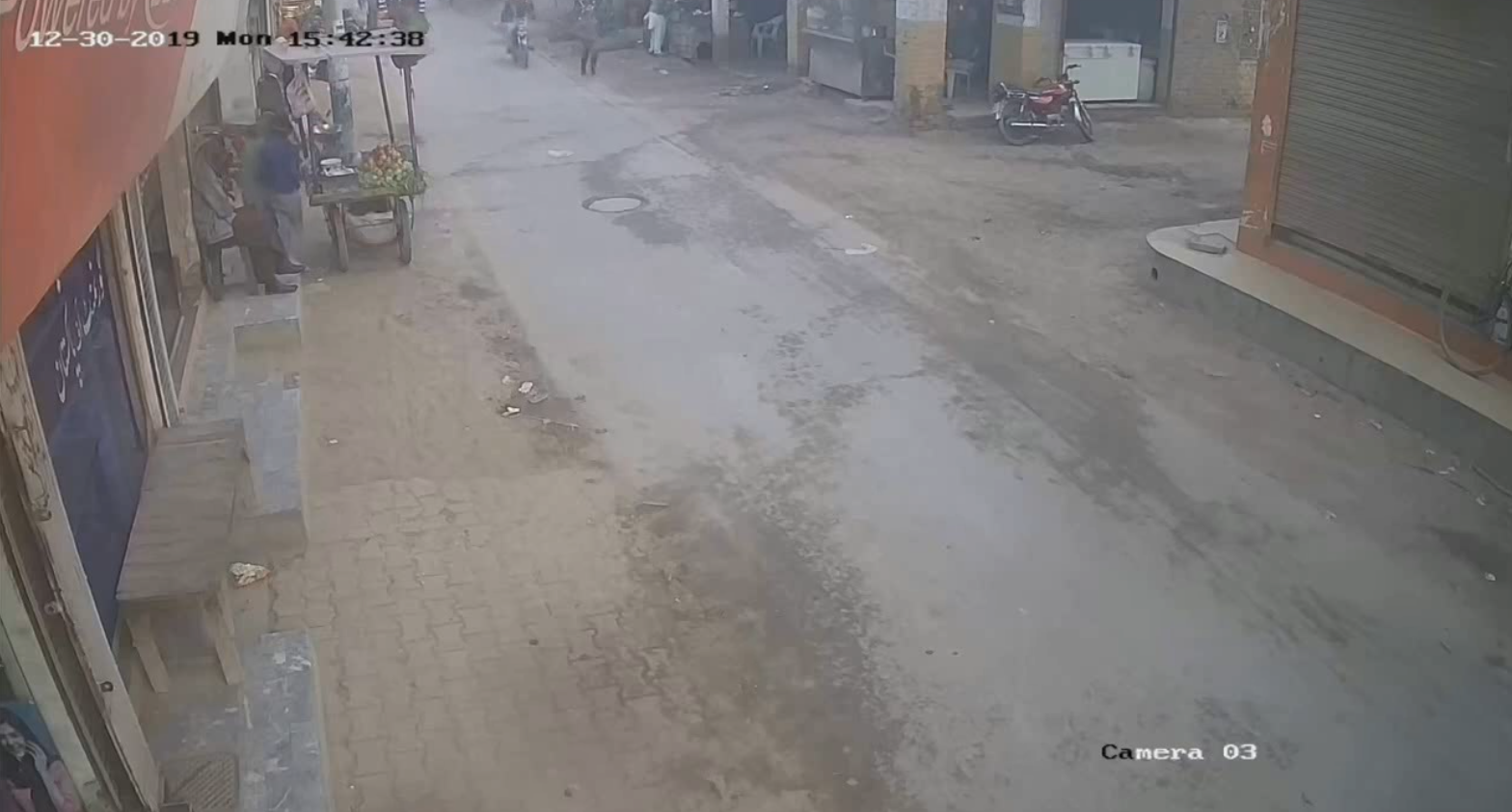}
    \hfill
    \includegraphics[width=0.31\columnwidth,height=2.6cm]{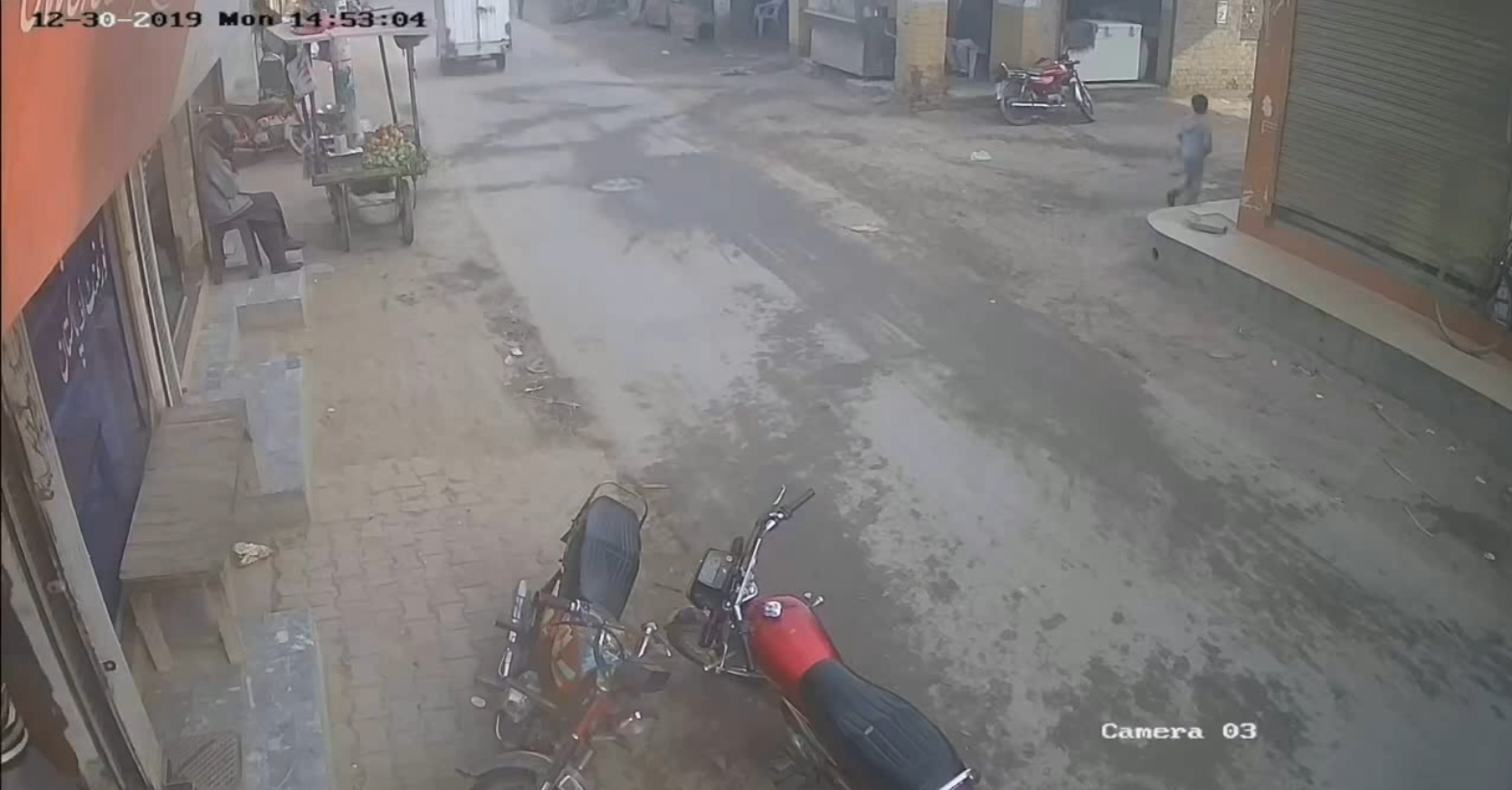}
    \hfill
    \includegraphics[width=0.31\columnwidth,height=2.6cm]{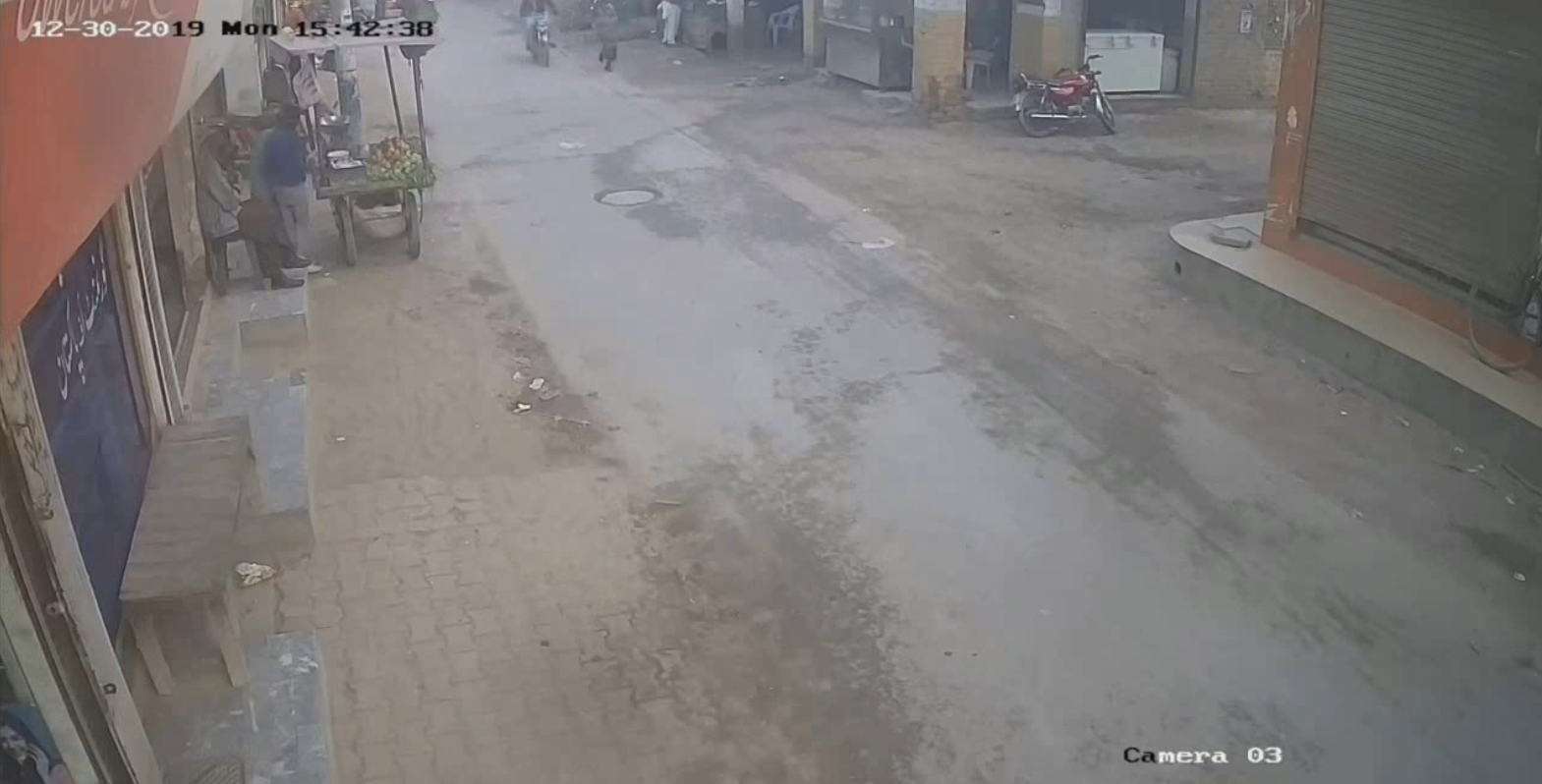}
    \caption{Example frames illustrating the CSVTED manipulation scenario. The dataset includes structured spatial and temporal tampering cases, particularly in surveillance-style videos, but it does not explicitly model short manipulated intervals inserted into otherwise authentic video streams.}
    \label{fig:csvted_examples}
\end{figure}

CSVTED \cite{csvted} was developed for structured video tampering evaluation, particularly in surveillance-style scenarios. It includes copy-move and spatio-temporal manipulations. Although it is relevant for surveillance analysis, it does not directly address the exact manipulation pattern targeted in our work. As illustrated in Figure~\ref{fig:csvted_examples}, CSVTED mainly focuses on structured spatial and temporal tampering patterns in surveillance-like settings rather than the insertion of a short manipulated interval into an otherwise authentic video stream.

\subsection{Speech Segment Localization and HAD}
Zhang and Sim \cite{zhang2022localizing} and HAD \cite{had} focus on audio-based partial manipulation detection and temporal localization of fake speech segments. While these studies are useful for analyzing temporally localized audio forgery, they are not suitable for our visual video manipulation scenario.

% TABLO 1: Veri setleri bitince, Limitations başlamadan önce burada.
\begin{table*}[htbp]
\caption{Comparison of existing datasets with respect to temporally localized manipulation scenarios}
\label{tab:dataset_comparison}
\centering
\resizebox{\textwidth}{!}{%
\begin{tabular}{lcccccc}
\toprule
Dataset & Year & Manipulation Type & Temporal Annotation & Object-Level Scene & Audio Modality & Inserted Fake Segment in Real Video \\
\midrule
LAV-DF \cite{lavdf}
& 2022
& Face / AV forgery
& Yes
& No
& Yes
& Partial \\

AV-Deepfake1M \cite{avdeepfake1m}
& 2024
& Face / speech manipulation
& Limited
& No
& Yes
& Partial \\

TVIL \cite{Zhang_2023}
& 2023
& Object removal / inpainting
& Yes
& Yes
& No
& No \\

DAVIS (basis of DAVIS-VI) \cite{davis2016}
& 2016
& Inpainting generation basis
& No
& Yes
& No
& No \\

VideoSham \cite{videosham}
& 2023
& Object insertion / removal
& Limited
& Yes
& No
& No \\

VPData / VPBench \cite{vpdata}
& 2025
& Inpainting / video editing
& No
& Yes
& No
& No \\

FSVD \cite{fsvd}
& 2023
& Object insertion / deletion
& No
& Yes
& No
& No \\

CSVTED \cite{csvted}
& 2025
& Copy-move / spatio-temporal
& Partial
& Yes
& No
& No \\

HAD \cite{had}
& 2021
& Word-level speech manipulation
& Yes
& No
& Yes
& Audio only \\

Fake Speech Segment Localization \cite{zhang2022localizing}
& 2022
& Fake speech segment localization
& Yes
& No
& Yes
& Audio only \\
\bottomrule
\end{tabular}
}
\end{table*}

As shown in Table~\ref{tab:dataset_comparison}, none of the existing datasets explicitly model the scenario where a short manipulated segment is inserted into an otherwise authentic video while preserving the temporal continuity of the original footage.

% ============================================================
\section{Proposed Dataset}
% ============================================================

In realistic tampering scenarios, the manipulated content may occupy 
only a very limited temporal interval while the remainder of the video 
remains authentic. This setting is challenging because the forged segment 
must be visually coherent with the original scene in terms of lighting, 
motion continuity, object placement, and temporal transitions. We first 
discuss the limitations of existing datasets, then describe our dataset 
construction methodology.

% ============================================================
\subsection{Limitations of Existing Datasets}
\label{sec:dataset_limitations}
% ============================================================

Although numerous datasets have been introduced for deepfake detection and video forgery analysis, most focus on facial deepfakes, speech synthesis, object removal, or fully manipulated scenes. Existing datasets rarely model scenarios in which a video remains mostly authentic, contains only a short manipulated segment, and then continues with the original authentic footage.

This type of manipulation is particularly challenging because the forged segment must preserve visual and temporal consistency with the surrounding authentic content. As a result, current benchmarks are insufficient for evaluating methods designed for temporally localized realistic manipulation detection in authentic video streams.

% ============================================================
\subsection{Dataset Construction}
\label{sec:proposed_dataset}
% ============================================================

To systematically evaluate the detection of temporally localized manipulations, we constructed a custom two-part dataset: a \textit{Pure Authentic Control Group} and a \textit{Manipulated (Merged) Set}.

\textbf{Source Material and Content:} The authentic videos were acquired through a combination of personal recordings and high-quality stock footage sourced from Pexels. The content was specifically curated to include both indoor environments (e.g., static office and home scenes) and outdoor nature landscapes. To ensure the control group captures natural temporal dynamics, the outdoor videos frequently feature transient background elements such as passing airplanes, flying birds, insects, falling leaves, or moving animals. This natural motion complexity is critical for evaluating whether a detection system can distinguish between genuine background dynamics and artificial manipulation boundaries.

\textbf{Manipulation Pipeline:} The creation of the partially manipulated videos followed a precise, programmatic pipeline to guarantee seamless temporal transitions. First, the 100th frame of each authentic video was systematically extracted using FFmpeg. Using this specific frame as a visual prompt and starting point, a short manipulated video segment was generated via advanced generative AI and video editing tools. Finally, a custom automated script was employed to splice the sequences together. The resulting manipulated videos follow a strict temporal structure: the sequence begins with the original authentic frames (frames 1 to 99), transitions into the inserted fake segment starting at frame 100, and seamlessly resumes the remaining authentic footage immediately after the fake segment concludes.

\begin{figure}[htbp]
    \centering
    % 1. Satır: Çiçek/Doğa Örnekleri (PNG) - Özgür bırakıldı (Orijinal oran)
    \includegraphics[width=0.32\columnwidth]{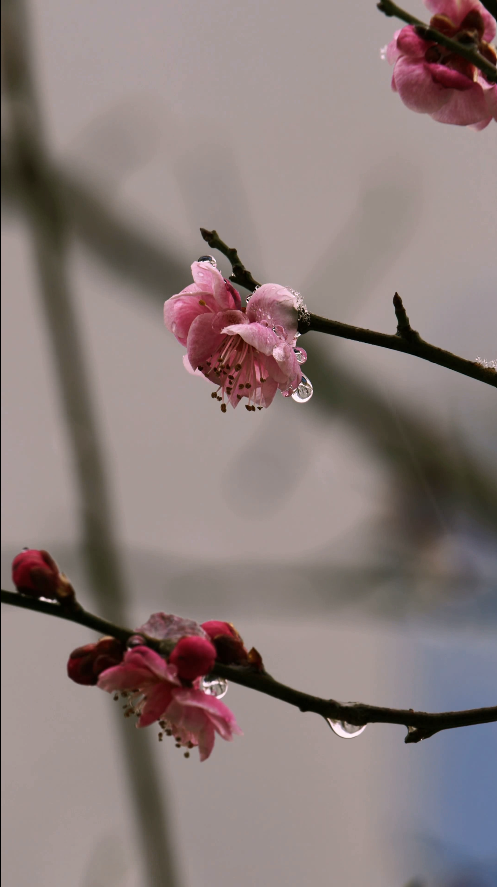}
    \hfill
    \includegraphics[width=0.32\columnwidth]{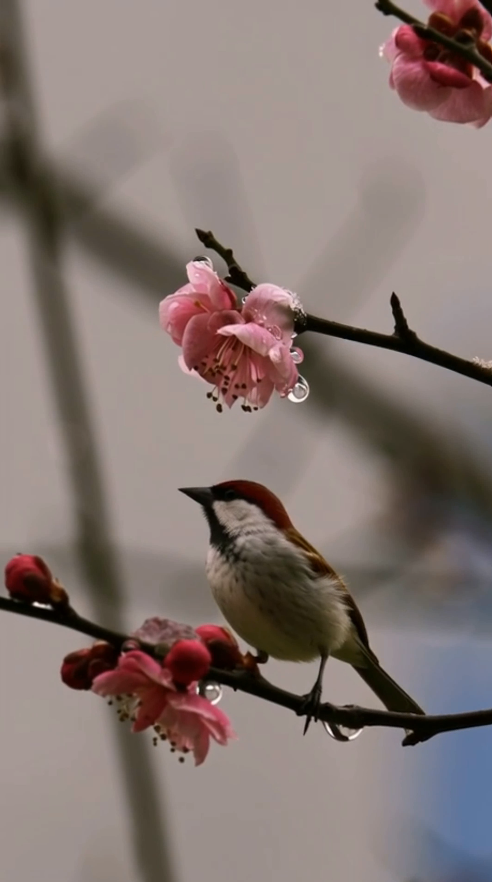}
    \hfill
    \includegraphics[width=0.32\columnwidth]{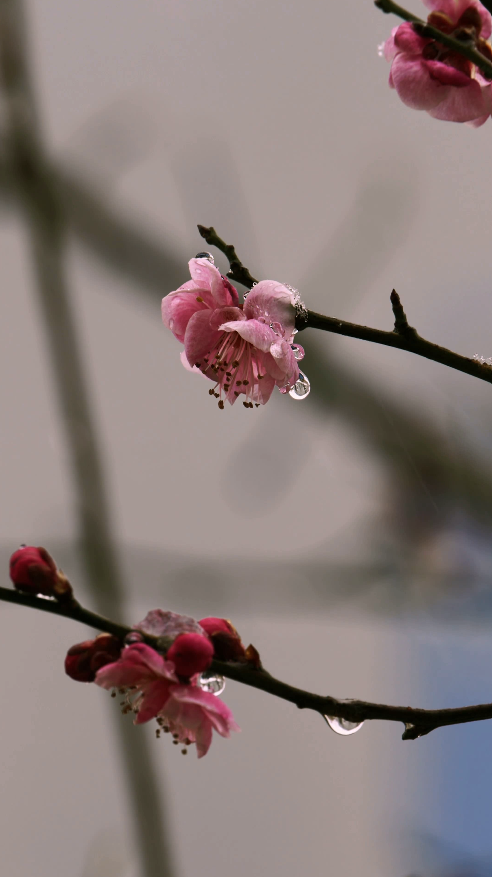}

    \vspace{1.5mm} % Satırlar arası dengeli boşluk

    % 2. Satır: Kedi Örnekleri (PNG) - Özgür bırakıldı (Orijinal oran)
    \includegraphics[width=0.32\columnwidth]{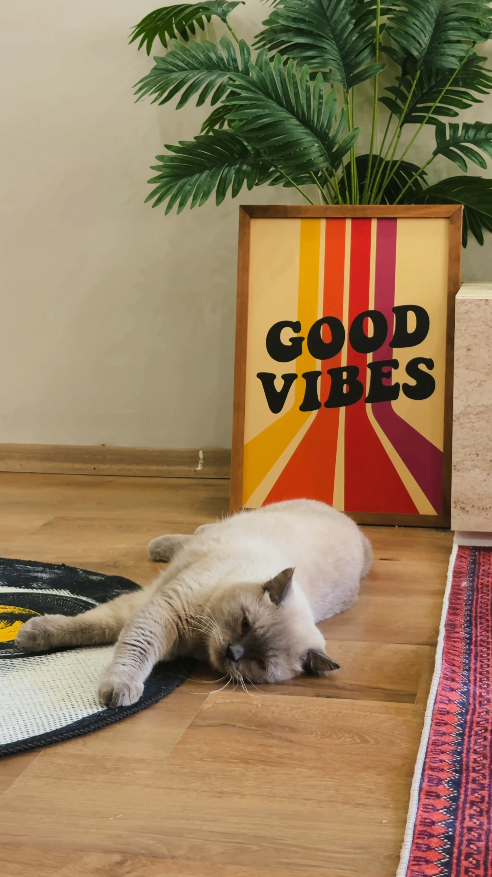}
    \hfill
    \includegraphics[width=0.32\columnwidth]{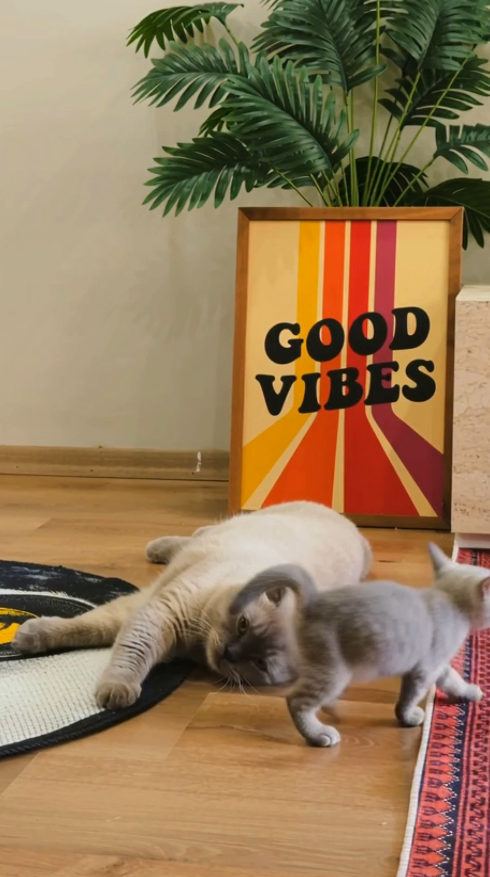}
    \hfill
    \includegraphics[width=0.32\columnwidth]{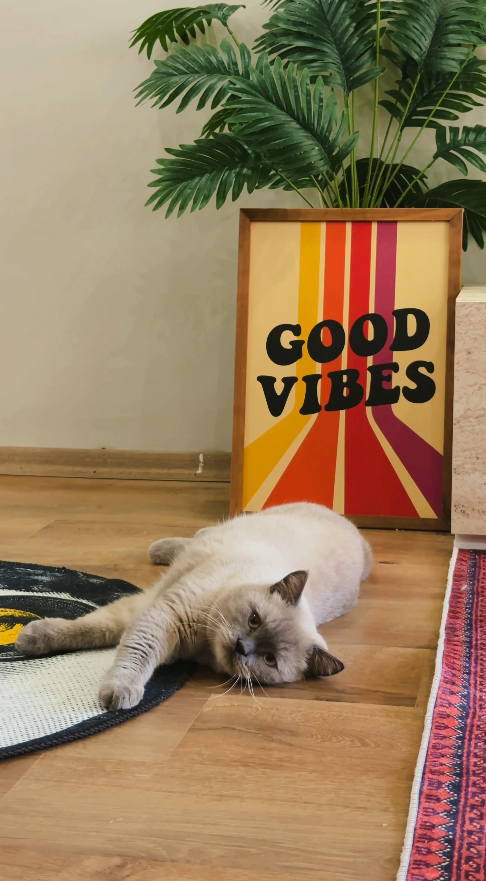}

    \vspace{1.5mm}

    % 3. Satır: Uçak Örnekleri (PNG) - Senin istediğin gibi sabit bırakıldı
    \includegraphics[width=0.32\columnwidth,height=2.0cm]{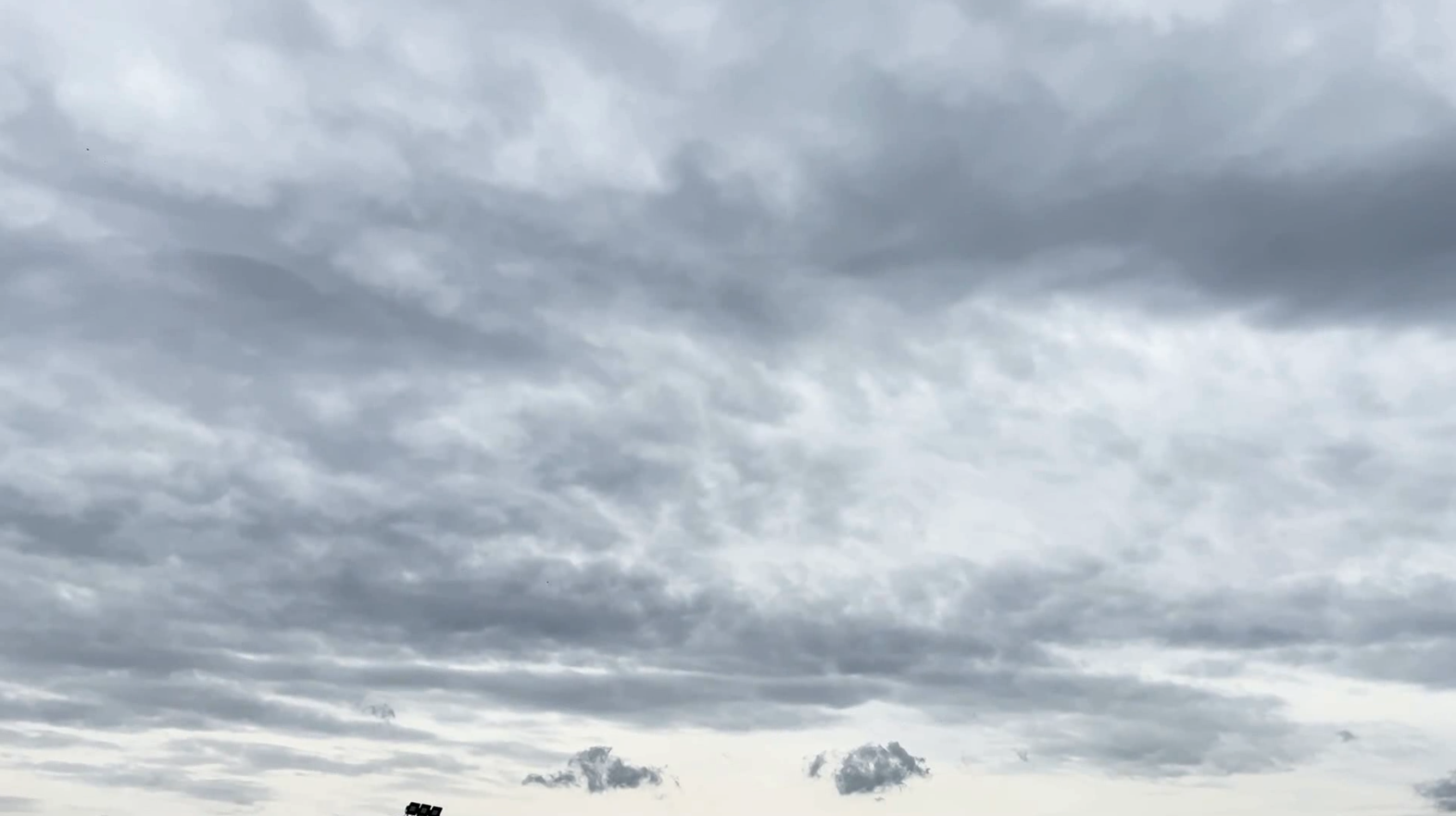}
    \hfill
    \includegraphics[width=0.32\columnwidth,height=2.0cm]{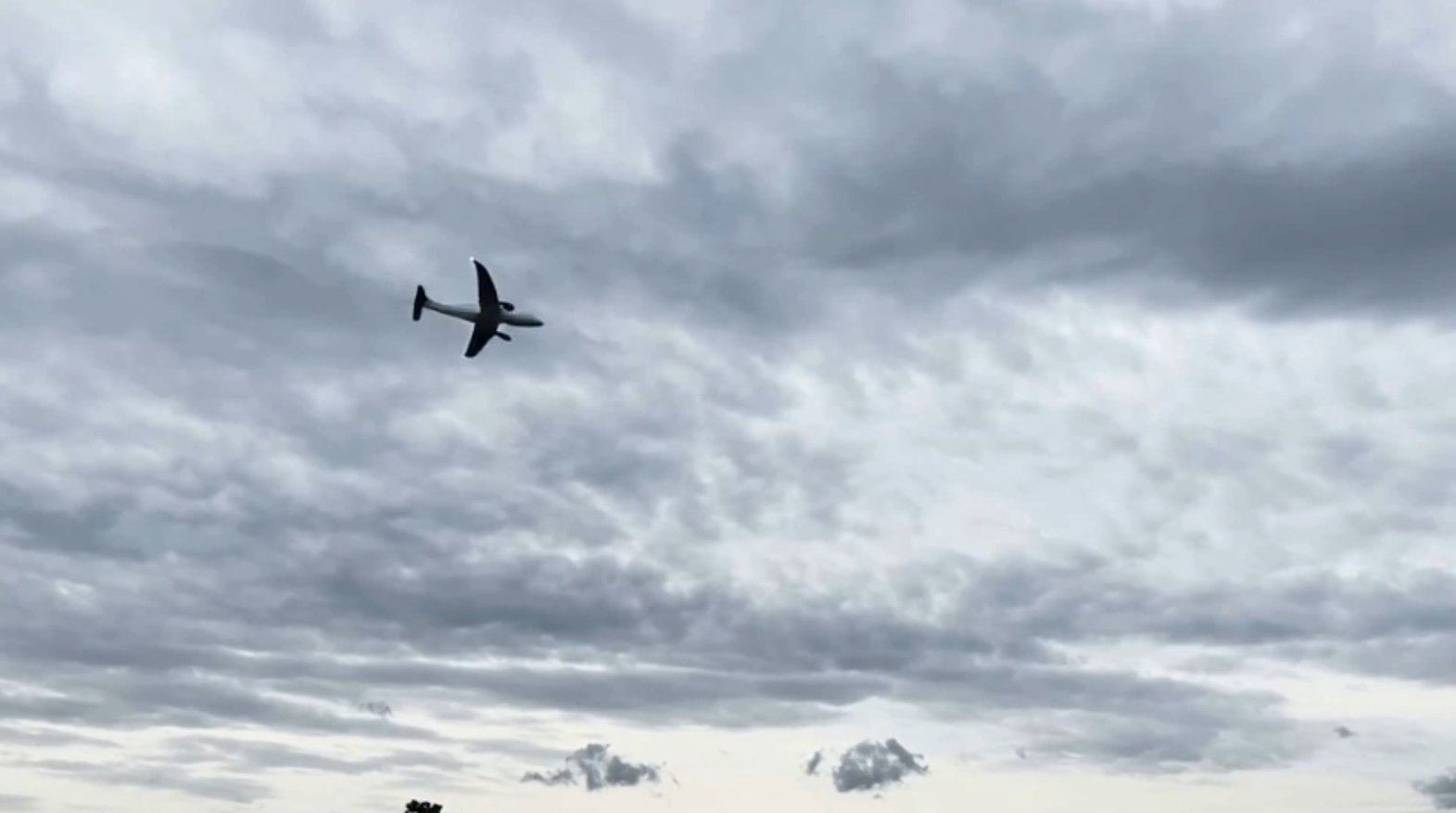}
    \hfill
    \includegraphics[width=0.32\columnwidth,height=2.0cm]{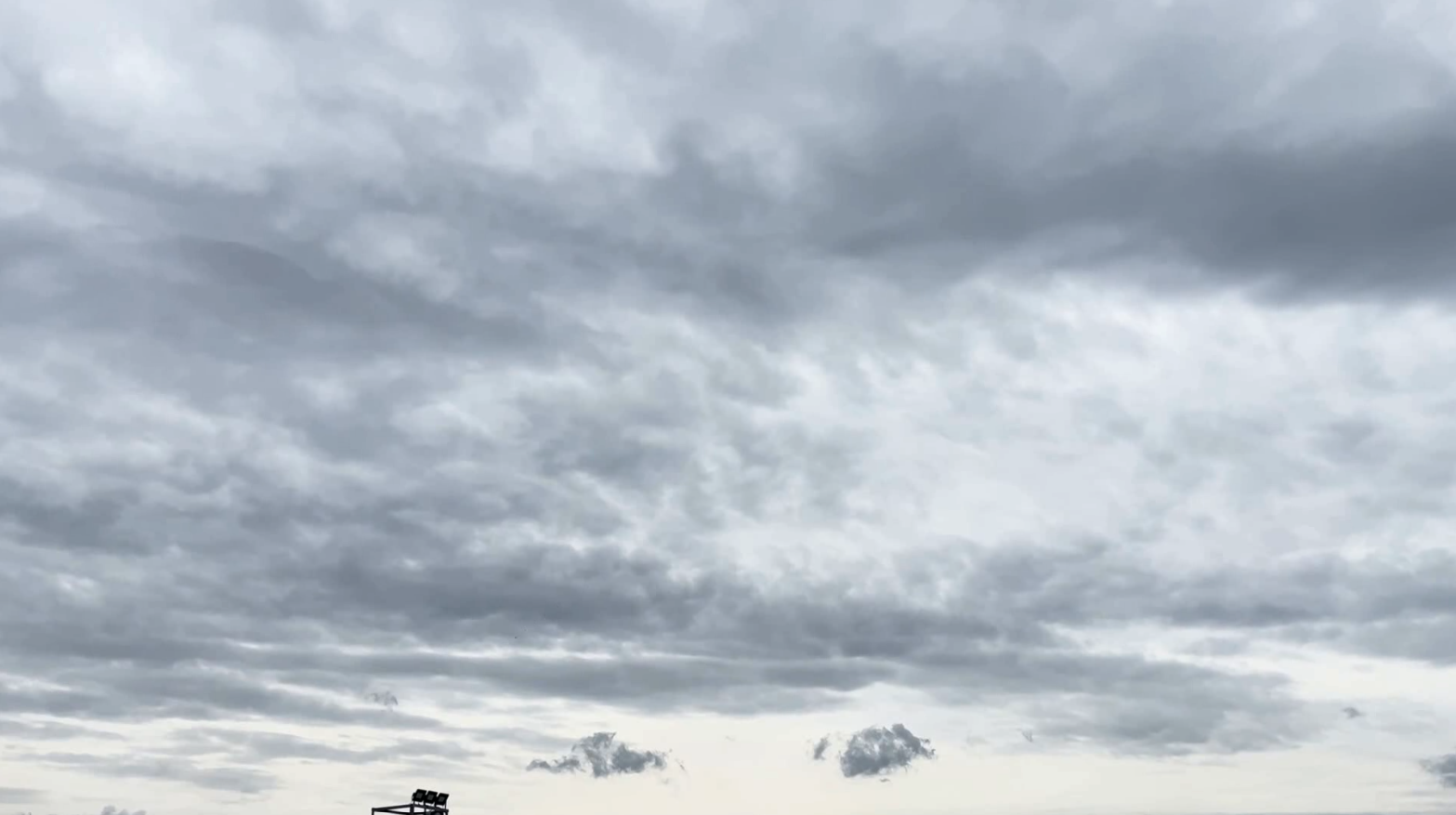}
    
    \caption{Representative samples from our custom-curated dataset. Rows demonstrate various natural background and foreground motions: (Top) natural elements such as flowers in the wind, (Middle) a kitten passing in front of a resting cat, and (Bottom) passing airplanes. This organic motion complexity challenges the model to differentiate between genuine environmental dynamics and artificial manipulation boundaries.}
    \label{fig:custom_dataset_examples}
\end{figure}

\textbf{Dataset Statistics:} The final dataset comprises 100 pure authentic videos, 100 generated manipulation segments, and 100 corresponding partially manipulated (merged) videos. The temporal properties of the videos vary to reflect diverse real-world conditions. Based on our insertion logs, the original frame rates range between 24.0 and 50.0 frames per second (FPS), with the vast majority recorded at 24.0 FPS. The lengths of the inserted fake segments vary strictly depending on the content (typically between 125 and 262 frames), resulting in final merged videos with diverse total durations ranging from approximately 300 to over 1300 frames. At the frame level, every sequence is densely annotated, indicating exact manipulation start and end boundaries. Representative samples from the dataset, illustrating the 
diversity of scene types and natural motion patterns, are 
shown in Figure~\ref{fig:custom_dataset_examples}.

\section{Methodology}

Our video manipulation detection framework explores two complementary approaches to evaluate model generalizability on temporally localized, realistic video manipulations. The first approach trains a linear classifier on top of frozen, pre-trained DINOv3 features using a diverse, multi-domain dataset. The second approach leverages DINOv3 features and a consecutive frame similarity-based method to capture temporal inconsistencies without relying on extensive forged training data. Both methods are tested on our newly curated, realistic video manipulation dataset to assess their effectiveness in challenging, partially manipulated scenarios. The following subsections detail each approach and the construction 
of the training dataset used for the supervised baseline.

\subsection{Detection Model: Linear Probe on DINOv3 Features}

We employ a linear classifier (linear probe) trained on top of pre-trained features from DINOv3~\cite{simeoni2025dinov3}, a self-supervised 
Vision Transformer (ViT). We utilize the ViT-Base variant of DINOv3, which processes images at a standard resolution, producing a high-dimensional feature vector from its token outputs. Specifically, we extract the token-level representations from the backbone and apply pooling to obtain a compact feature representation. A single linear layer is then attached to this frozen feature extractor, which is trained to classify the pooled features as real or fake. This approach serves as a strong baseline, evaluating the generalizability of features learned through large-scale self-supervised pre-training without fine-tuning the backbone.

To train a model capable of generalizing beyond specific manipulation techniques, we deliberately constructed a training dataset from multiple independent sources, ensuring no overlap with our evaluation set.

\textbf{Real Content:} Authentic videos and images were sourced from publicly available stock footage repositories, providing diverse content across various scenes, lighting conditions, and subjects. Additionally, we included real facial videos from publicly available face datasets to ensure the model learns authentic facial texture and motion dynamics.

\textbf{Fake Content:} Artificially generated images were primarily sourced from the DALL-E Recognition Dataset \cite{nathan_koliha_2024}, which contains images generated by AI models such as DALL-E and Midjourney alongside human-made art. This introduces a wide variety of generative artifacts, from diffusion-based noise patterns to structural inconsistencies. We also incorporated fake facial videos generated by various deepfake manipulation techniques, including face-swapping and reenactment methods, to expose the model to both spatial and temporal forgery patterns common in video deepfakes.

By mixing these different domains — real stock footage, authentic faces, AI-generated art, and manipulated facial videos — our training set challenges the model to learn domain-agnostic forgery cues rather than dataset-specific biases. Importantly, this training collection is completely disjoint from our evaluation dataset, ensuring a fair assessment of cross-domain generalization. The model was implemented in PyTorch, trained with a frozen DINOv3 backbone, optimizing only the final linear layer.

\subsection{DINOv3 Feature Similarity Approach}

Our second approach employs a training-free, feature-based method for detecting temporally localized video manipulations. This technique leverages DINOv3, a self-supervised vision transformer, to extract high-level semantic features from each video frame. The core intuition is that manipulated segments introduce discontinuities in the visual feature space, which can be detected by analyzing the similarity between consecutive frames.

For a given video, we extract DINOv3 feature vectors for every frame and compute the cosine distance between each consecutive pair. This yields a temporal distance signal where abrupt changes indicate potential manipulation boundaries. To adapt to local variations in motion and scene complexity, we compute a rolling window-based Z-score over the distance sequence. Specifically, for each frame, we calculate the local mean and standard deviation over a window of size \(P\) frames, then derive the Z-score as:

\[
Z_t = \frac{d_t - \mu_t}{\sigma_t + \epsilon}
\]

where \(d_t\) is the cosine distance at frame \(t\), \(\mu_t\) and \(\sigma_t\) are the local mean and standard deviation, and \(\epsilon\) is a small stabilizer term.

Manipulated intervals are identified as contiguous regions where the Z-score exceeds a predefined threshold \(\tau\) and the raw distance surpasses a minimum sensitivity threshold \(\delta\). The start of the forged segment is taken as the earliest frame meeting these criteria, and the end as the latest such frame. This produces a continuous predicted manipulation interval, which we evaluate against ground truth annotations using standard classification metrics. The method requires no training or fine-tuning, making it highly generalizable across unseen manipulation types and datasets.

Algorithm~\ref{alg:dinov3_detection} summarizes the complete detection procedure.

\begin{algorithm}[htbp]
\caption{DINOv3-Based Temporal Manipulation Detection}
\label{alg:dinov3_detection}
\begin{algorithmic}[1]
\REQUIRE Video frames $F_1, F_2, \ldots, F_N$, window size $P$, Z-score threshold $\tau$, distance threshold $\delta$, stabilizer $\epsilon$
\ENSURE Predicted manipulation labels $y_1, y_2, \ldots, y_N \in \{0,1\}$
\STATE Extract DINOv3 features: $\mathbf{v}_i = \text{DINOv3}(F_i)$ for $i = 1..N$
\STATE Compute cosine distances: $d_i = 1 - \frac{\mathbf{v}_i \cdot \mathbf{v}_{i+1}}{\|\mathbf{v}_i\|\|\mathbf{v}_{i+1}\|}$ for $i = 1..N-1$
\FOR{$t = 1$ to $N-1$}
    \STATE Compute $\mu_t = \text{mean}(\{d_{t-P/2}, \ldots, d_{t+P/2}\})$
    \STATE Compute $\sigma_t = \text{std}(\{d_{t-P/2}, \ldots, d_{t+P/2}\})$
    \STATE $Z_t = (d_t - \mu_t) / (\sigma_t + \epsilon)$
\ENDFOR
\STATE Identify candidate frames: $\mathcal{C} = \{t \mid Z_t > \tau \;\text{and}\; d_t > \delta\}$
\IF{$|\mathcal{C}| \ge 2$}
    \STATE $start = \min(\mathcal{C})$, $end = \max(\mathcal{C})$
    \STATE Set $y_t = 1$ for all $t \in [start, end]$
\ELSE
    \STATE Set $y_t = 0$ for all $t$
\ENDIF
\RETURN $\{y_t\}_{t=1}^{N-1}$
\end{algorithmic}
\end{algorithm}

% ============================================================
\section{Experimental Results}
% ============================================================
All methods are evaluated using frame-level binary classification metrics. 
In this context, a True Positive (TP) is a frame predicted as fake that is 
genuinely manipulated; a False Positive (FP) is a real frame incorrectly 
predicted as fake; and a False Negative (FN) is a manipulated frame 
incorrectly predicted as real. Precision, recall, and F1-score are derived 
from these counts as follows: Precision $= \text{TP}/(\text{TP}+\text{FP})$, 
Recall $= \text{TP}/(\text{TP}+\text{FN})$, and 
F1 $= 2 \cdot \text{Precision} \cdot \text{Recall} / (\text{Precision} + \text{Recall})$.

\subsection{Performance of the Linear Probe Baseline}
We evaluated the Linear Probe on DINOv3 features using three detection strategies on our custom-curated, partially manipulated video test set. The evaluation process is as follows:

\begin{enumerate}
    \item The model processes all test videos frame-by-frame.
    \item For each frame, the model outputs a probability score $p \in [0,1]$ indicating the likelihood of the frame being fake.
    \item We compare three thresholding strategies to convert these probabilities into binary decisions:
        \begin{itemize}
            \item \textbf{Fixed threshold:} A standard threshold of 0.5 is applied uniformly across all frames.
            \item \textbf{Video-level adaptive threshold:} For each video, we compute the mean ($\mu$) and standard deviation ($\sigma$) of the predicted fake probabilities across all its frames. An adaptive threshold is then set at $\mu + 1.5\sigma$, classifying frames with a probability exceeding this value as fake. This technique assumes that fake frames represent statistical anomalies within the video stream, making them detectable as outliers.
            \item \textbf{Sliding window threshold:} A temporal sliding window of odd size $W$ (e.g., 21 frames) is applied over consecutive frames. For each frame, we examine its local neighborhood defined by the window centered on that frame. Within this window, we count how many frames would be classified as fake using a fixed threshold of 0.5. If the ratio of such frames within the window meets or exceeds a predefined threshold $T$ (e.g., 0.6), the entire window — including the center frame — is classified as fake. Otherwise, the center frame is classified as real. This approach propagates high-confidence fake detections to surrounding frames, effectively filling short temporal gaps and reducing false negatives in partially manipulated segments.
        \end{itemize}
    \item The resulting binary predictions are compared against the frame-level ground truth to compute per-video and aggregate metrics including precision, recall, and F1-score.
\end{enumerate}

This testbed specifically targets a challenging real-world scenario — partial video manipulation — and allows us to measure how well a model trained on disparate, primarily static image artifacts can generalize to detecting manipulations in dynamic video content where fake segments are inserted into otherwise authentic sequences.

\begin{table}[htbp]
\centering
\caption{Linear Probe Performance Across Detection Strategies}
\label{tab:linear_probe}
\begin{tabular}{|l|c|c|c|}
\hline
\textbf{Method} & \textbf{Precision} & \textbf{Recall} & \textbf{F1-score} \\
\hline
Default        & 0.311 & 0.223 & 0.260 \\
Adaptive       & \textbf{0.473} & 0.098 & 0.162 \\
Sliding Window ($W$=21, $T$=0.6) & 0.245 & 0.217 & 0.230 \\
\hline
\end{tabular}
\end{table}

\subsubsection*{Evaluation Summary}

The experimental results reveal several critical findings as shared in 
Table~\ref{tab:linear_probe}. First, \textbf{we prioritize precision as 
the primary evaluation metric} for partial video deepfake detection in 
real-world deployment. In this scenario, false positives (classifying 
real frames as fake) erode user trust more rapidly than missed detections. 
A system that frequently raises false alarms will be abandoned by users, 
whereas occasional missed fakes may go unnoticed. This prioritization is 
consistent with established practice in multimedia forensics, where 
minimizing false alarms is critical to maintaining system credibility.

Second, among the three strategies, the \textbf{Adaptive threshold method achieves the highest precision (0.473)}. While its recall is low (0.098), meaning it detects only 9.8\% of fake frames, this performance can be acceptable when considering the dataset profile. Given that fake frames constitute only 29\% of the entire dataset, a random classifier would yield a precision of approximately 0.29.This result is further supported by the following observations: 

\begin{itemize}
    \item When the model flags a frame as fake, it is correct 47.3\% of the time — substantially better than random guessing or the Default strategy.
    \item The adaptive threshold ($\mu + 1.5\sigma$) successfully filters out the majority of false positives by requiring detected anomalies to be statistically significant outliers within each video's probability distribution.
    \item In practice, a conservative detector with high precision can be integrated into human-in-the-loop systems, where flagged frames undergo manual verification.
\end{itemize}

Although the Adaptive strategy achieves the highest precision, its very 
low recall (0.098) results in the lowest F1-score (0.162) among the three 
strategies, reflecting the inherent precision-recall trade-off in this 
conservative detection approach.

Third, the \textbf{Default threshold (0.5) and Sliding Window strategies 
yield lower precision} (0.311 and 0.245 respectively) despite higher recall. This reflects the challenge of partial video manipulation: the model's raw probability outputs are not well-calibrated, and simple thresholds or local window averaging fail to distinguish authentic frames from manipulated ones with sufficient specificity.

Finally, the \textbf{low overall performance across all strategies} (maximum precision 0.473) highlights the difficulty of cross-domain generalization. The linear probe, trained primarily on static images (DALL-E generated art, face-swapped images), struggles to generalize to partially manipulated videos where fake segments are brief and embedded within long authentic sequences. Future work should explore:
\begin{itemize}
    \item Temporal modeling (e.g., LSTM or Transformer layers on top of frame-level features)
    \item Video-specific pre-training objectives
    \item Calibration techniques to improve probability estimates
    \item Class-balanced losses to address the extreme class imbalance in partial video manipulation
\end{itemize}

\begin{figure*}[htbp]
\centering

\begin{minipage}{\textwidth}
    \centering
    \includegraphics[width=0.85\textwidth]{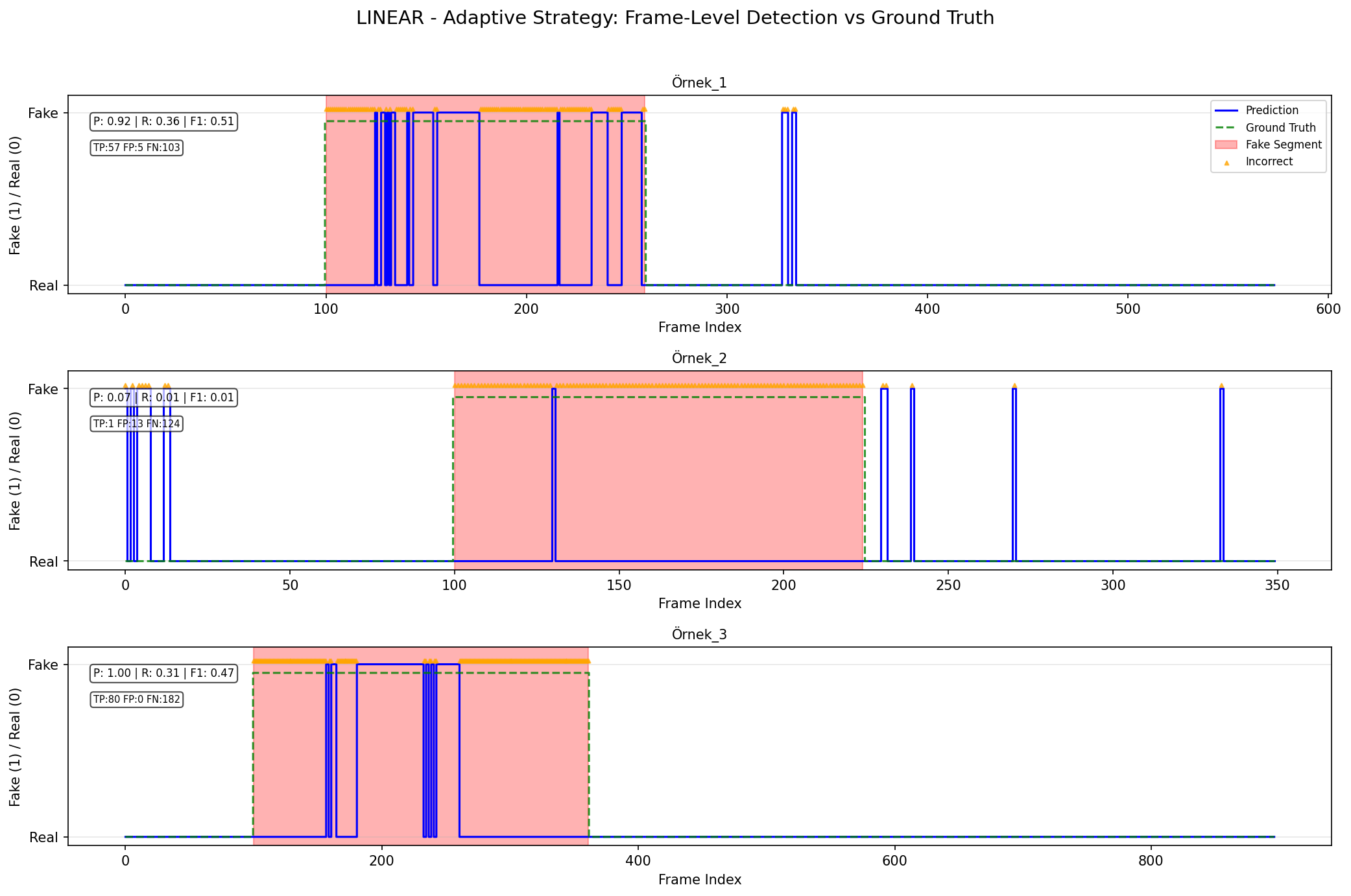}
    \captionof{figure}{Frame-level detection visualization for the first three test videos using the Linear Probe with the Adaptive threshold strategy. Blue steps represent the model's binary predictions (1~=~fake, 0~=~real). Green dashed steps show the ground truth labels. Red shaded regions indicate contiguous fake segments in the ground truth. Orange triangles mark individual frames where the prediction differs from the ground truth. Performance metrics (precision, recall, F1-score) are displayed for each video. The Adaptive strategy ($\mu + 1.5\sigma$) produces very few positive predictions (visible as isolated blue steps). This conservative approach prioritizes trustworthiness over coverage, making it more suitable for real-world deployment where false alarms are costly.}
    \label{fig:detection_examples_1}
\end{minipage}

\vspace{6mm}

\begin{minipage}{\textwidth}
    \centering
    \includegraphics[width=0.85\textwidth]{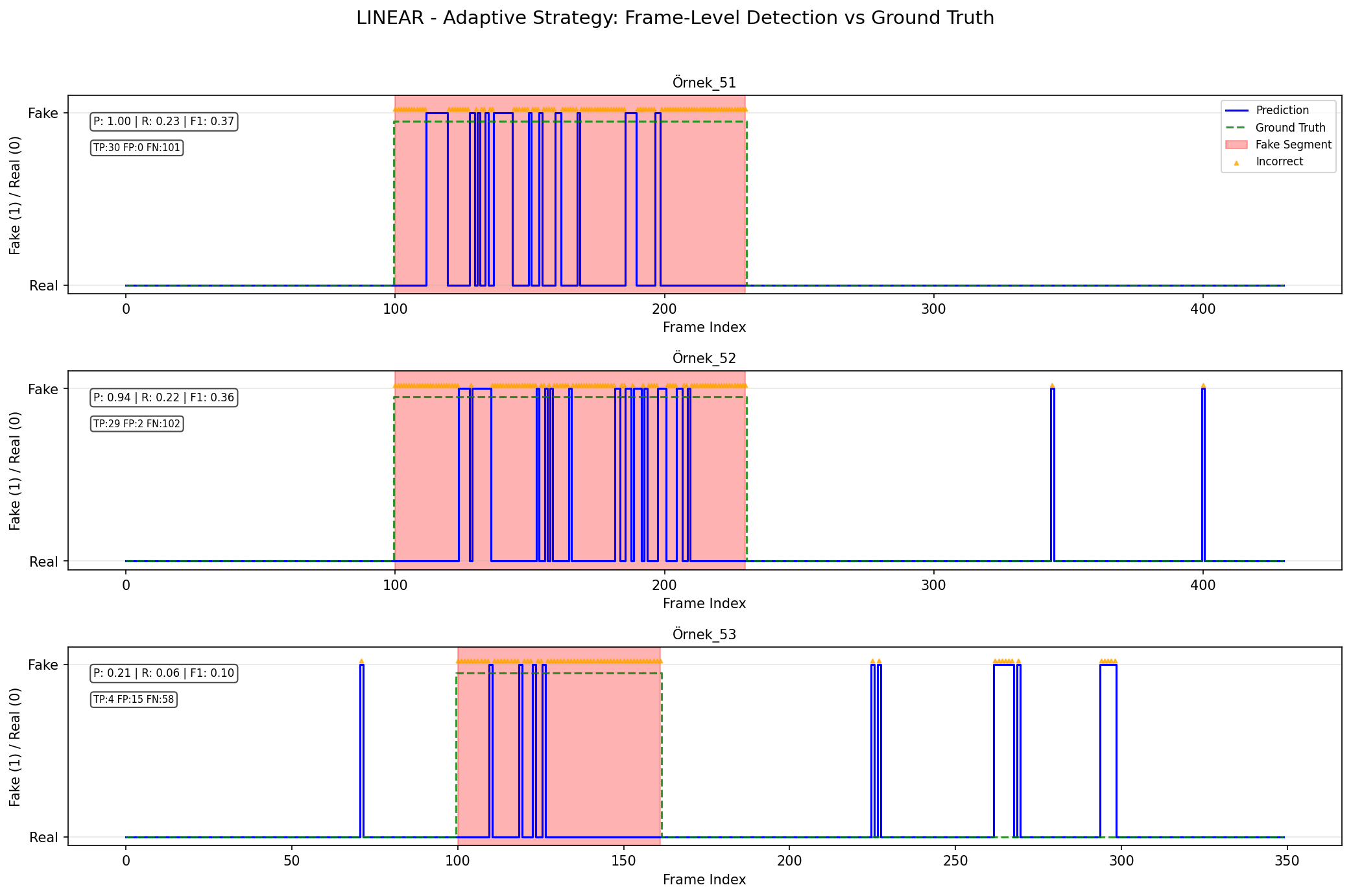}
    \captionof{figure}{Frame-level detection visualization for videos 50--53 using the Linear Probe with the Adaptive threshold strategy.}
    \label{fig:detection_examples_2}
\end{minipage}

\end{figure*}

Figure~\ref{fig:detection_examples_1} and \ref{fig:detection_examples_2} visualize the frame-level predictions for example videos using the Adaptive strategy. The figures show that the Linear Probe produces extremely sparse positive predictions — only a handful of frames across all videos are classified as fake. This explains the low recall (0.098): most fake frames (red shaded regions) are not detected. However, the orange triangles (incorrect predictions) are also sparse, and most positive predictions align with actual fake segments. This confirms that the Adaptive threshold successfully identifies only the most statistically anomalous frames, prioritizing precision over recall. While this approach is not yet production-ready, it demonstrates that DINOv3 features contain \emph{some} signal for deepfake detection, and future work should focus on improving sensitivity without sacrificing precision.

% BÜYÜK HİLE: İki resmi tek bir figure* içine aldık.
% Artık ayrılmaları, farklı sayfalara uçmaları veya References sonrasına gitmeleri imkansız.
% Sayfada tam Conclusion'dan önce alt alta basılacaklar.

\subsection{Performance of the Proposed Feature Similarity Approach}

To demonstrate the critical importance of temporal context, we evaluated our unsupervised DINOv3 Feature Similarity framework on the same evaluation dataset. Unlike the frame-by-frame baseline (Linear Probe), this approach operates directly on the temporal cosine distance between consecutive frames, allowing it to capture abrupt structural disruptions in the feature space.

The detailed global performance metrics extracted from the entire evaluation set are reported in Table~\ref{tab:proposed_performance}. Notably, the proposed framework achieves a high Global Precision of 83.00\%. In multimedia forensics, precision is often prioritized over recall, as maintaining user trust requires ensuring that flagged anomalies are genuinely manipulated rather than natural artifacts.

\begin{table}[htbp]
\centering
\caption{Global Frame-Level Performance of the Proposed Feature Similarity 
Approach. The merged set contains both authentic and manipulated frames from 
the same source videos; total frame counts differ from the control group due 
to the inclusion of inserted fake segments.}
\label{tab:proposed_performance}
\begin{tabular}{|l|c|}
\hline
\textbf{Metric} & \textbf{Value} \\
\hline
Total Processed Frames & 52,889 \\
Total Actual Fake Frames & 15,566 \\
Total Detected Fake Frames & 10,059 \\
\hline
Global Accuracy  & 83.12\% \\
Global Precision & 83.00\% \\
Global Recall    & 53.64\% \\
Global F1-score  & 65.16\% \\
\hline
\end{tabular}
\end{table}

\begin{figure}[htbp]
    \centering
    \includegraphics[width=\columnwidth]{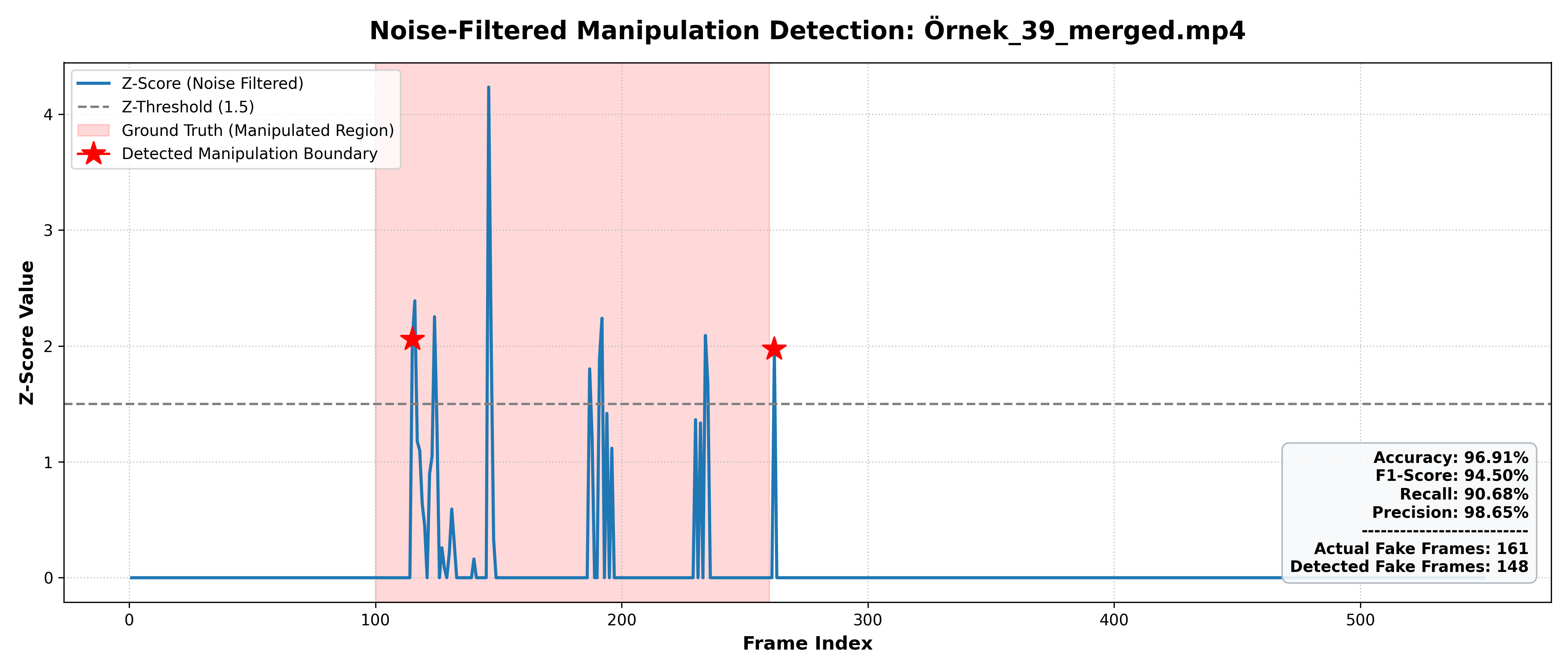}
    \vspace{1mm}
    \parbox{\columnwidth}{\centering (a) Successful Localization (Video 39)\\
    {[}Actual Fake Frames: 161, Detected Fake Frames: 148{]}}
    \vspace{4mm}
    
    \includegraphics[width=\columnwidth]{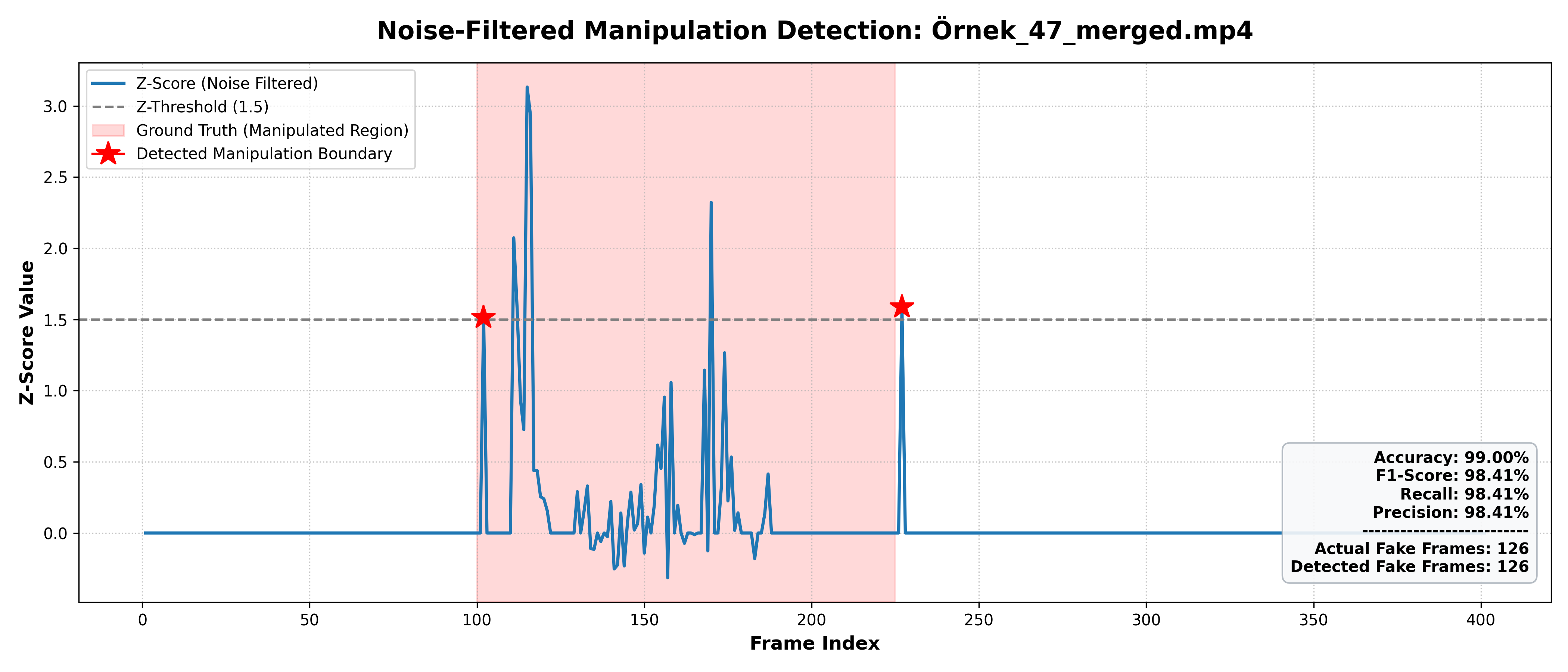}
    \vspace{1mm}
    \parbox{\columnwidth}{\centering (b) Successful Localization (Video 47)\\
    {[}Actual Fake Frames: 126, Detected Fake Frames: 126{]}}
    \vspace{4mm}
    
    \includegraphics[width=\columnwidth]{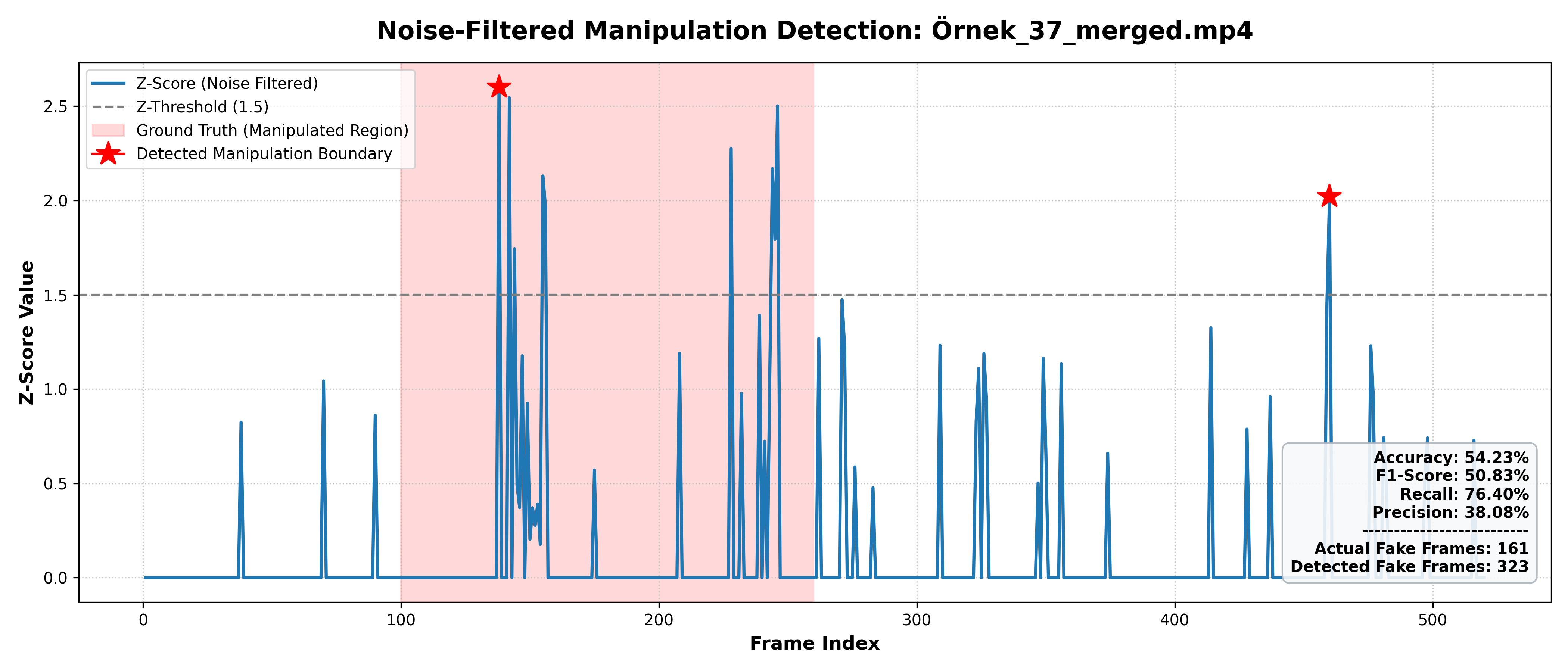}
    \vspace{1mm}
    \parbox{\columnwidth}{\centering (c) Imperfect Detection — Over-Extension (Video 37)\\
    {[}Actual Fake Frames: 161, Detected Fake Frames: 323{]}}
    
    \caption{Qualitative detection trajectories of the proposed Feature Similarity framework on the partially manipulated dataset. Shaded areas represent the ground-truth manipulation intervals, while red stars indicate the detected boundaries.}
    \label{fig:merged_analysis_plots}
\end{figure}

To provide a transparent qualitative analysis, Figure~\ref{fig:merged_analysis_plots} visualizes the frame-level detection trajectories for three representative cases from the manipulated set. Figure~\ref{fig:merged_analysis_plots}(a) and Figure \ref{fig:merged_analysis_plots}(b) demonstrate successful detection scenarios where the integration of the minimum sensitivity threshold ($\delta = 0.01$) successfully flattens background micro-movements, allowing the Z-score peaks to perfectly isolate the precise manipulation boundaries (Examples 39 and 47). 

Conversely, Figure \ref{fig:merged_analysis_plots}(c) illustrates an imperfect 
detection case involving temporal over-extension (Example 37). In this scenario, 
the Z-score signal remains elevated well beyond the true manipulation boundaries, 
causing the predicted interval to span 323 frames despite only 161 ground-truth 
fake frames. This over-detection likely arises because the transition back from 
the forged segment to the authentic footage produces a secondary distance spike 
that also surpasses the fixed threshold ($\tau = 1.5$), expanding the predicted 
interval beyond its true extent. This highlights a clear edge-case limitation of 
fixed global thresholding, where a single uniform threshold cannot distinguish 
genuine manipulation boundaries from post-transition settling artifacts.

\subsection{Robustness Evaluation on Pure Authentic Control Group}

A reliable multimedia forensics system must maintain an exceptionally low false-alarm rate when processing entirely authentic streams. To validate this robustness, we evaluated the proposed framework on the 100-video Pure Authentic Control Group under identical configuration parameters ($\tau = 1.5, \delta = 0.01$).

The comprehensive evaluation outcomes are detailed in Table \ref{tab:control_performance}. At the video level, the unsupervised framework achieved an exceptional robustness rate, triggering zero false alarms in 95 out of 100 entirely authentic videos, yielding a video-level accuracy of 95.00\%. Furthermore, out of 37,423 processed frames, 36,558 were correctly verified as True Negatives, resulting in a highly robust frame-level accuracy of 97.69\%.

\begin{table}[htbp]
\centering
\caption{Robustness Evaluation on the Pure Authentic Control Group}
\label{tab:control_performance}
\begin{tabular}{|l|c|}
\hline
\textbf{Metric} & \textbf{Value} \\
\hline
\multicolumn{2}{|c|}{\textbf{Video-Level Analysis}} \\
\hline
Total Tested Videos & 100 \\
Videos with Zero Errors & 95 \\
Videos with False Alarms & 5 \\
Video-Level Accuracy & 95.00\% \\
\hline
\multicolumn{2}{|c|}{\textbf{Frame-Level Analysis}} \\
\hline
Total Processed Frames & 37,423 \\
True Negative Frames & 36,558 \\
False Positive Frames & 865 \\
Frame-Level Accuracy & 97.69\% \\
\hline
\end{tabular}
\end{table}

\begin{figure}[htbp]
    \centering
    \includegraphics[width=\columnwidth]{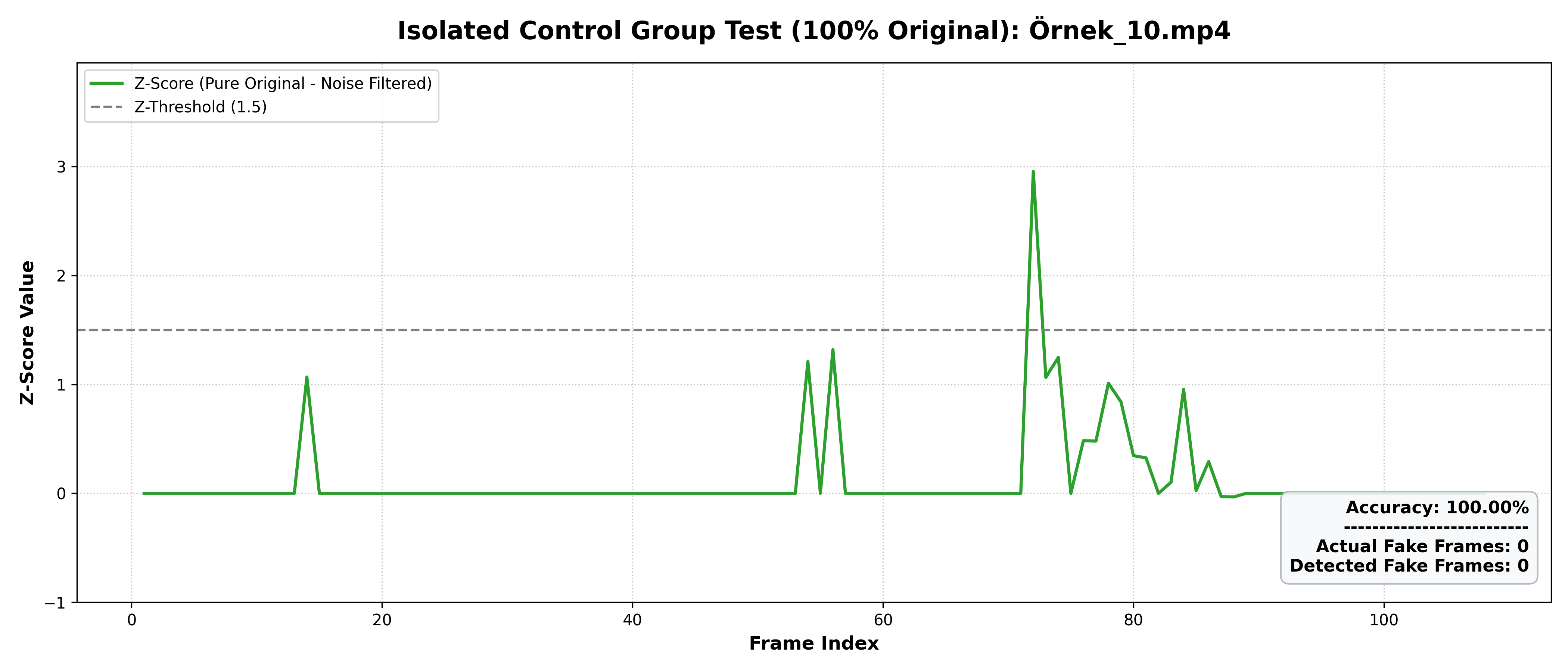}
    \vspace{1mm}
    \parbox{\columnwidth}{\centering (a) True Negative - Perfect Stability (Video 10)\\
    {[}Actual Fake Frames: 0, Detected Fake Frames: 0{]}}
    \vspace{4mm}
    
    \includegraphics[width=\columnwidth]{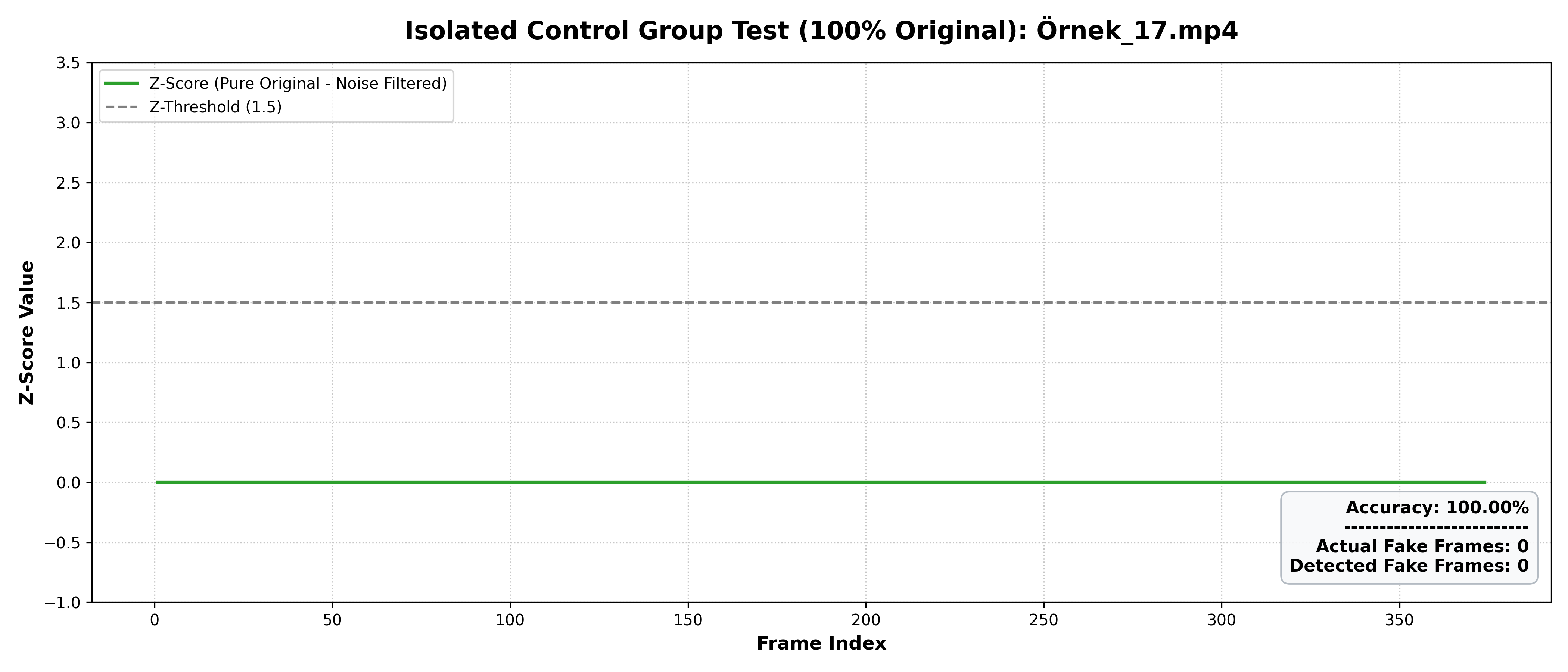}
    \vspace{1mm}
    \parbox{\columnwidth}{\centering (b) True Negative-Complex Motion Filtered (Video 17)\\
    {[}Actual Fake Frames: 0, Detected Fake Frames: 0{]}}
    \vspace{4mm}
    
    \includegraphics[width=\columnwidth]{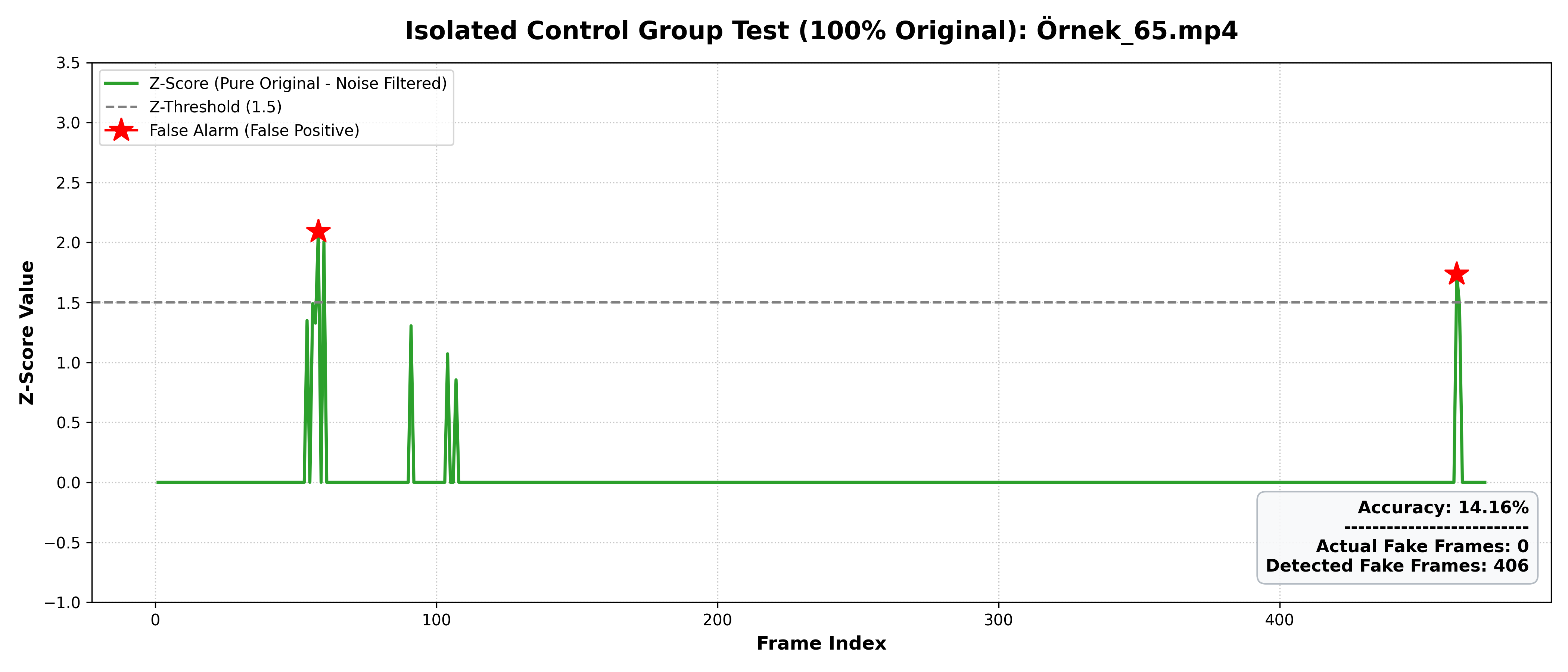}
    \vspace{1mm}
    \parbox{\columnwidth}{\centering (c) False Alarm Case (Video 65)\\
    {[}Actual Fake Frames: 0, Detected Fake Frames: 406{]}}
    
    \caption{Qualitative trajectories on the Pure Authentic Control Group. The proposed framework successfully maintains absolute stability across 95\% of unmanipulated streams, with rare false positives triggered only by extreme environmental lighting or motion spikes.}
    \label{fig:control_analysis_plots}
\end{figure}

Figure~\ref{fig:control_analysis_plots} illustrates the qualitative Z-score trajectories for the control group. Figure~\ref{fig:control_analysis_plots}(a) and Figure \ref{fig:control_analysis_plots}(b) represent the successful majority (Videos 10 and 17), where the minimum sensitivity threshold perfectly neutralizes complex environmental motion (such as passing airplanes or moving animals), keeping the Z-score well below the operational threshold. 

On the other hand, Figure \ref{fig:control_analysis_plots}(c) isolates a false alarm scenario (Video 65). In this specific video, an abrupt change in local illumination or a sudden camera jerk produces a sudden spike that passes the minimum sensitivity threshold and surpasses the $\tau = 1.5$ threshold, triggering an erroneous manipulation flag. This specific qualitative failure directly underscores the necessity for 
content-adaptive thresholding mechanisms discussed in Section~\ref{sec:limitations}.

\section{Limitations}
\label{sec:limitations}

While the proposed DINOv3-based similarity approach demonstrates strong performance in detecting temporally localized manipulations, it has several limitations that warrant further investigation.

The most significant limitation concerns the use of fixed threshold parameters across all videos. In our implementation, we applied uniform values for the Z-score threshold \(\tau = 1.5\), the minimum distance threshold \(\delta = 0.01\), and the window size \(P = 30\) to every video in the test set. However, different video characteristics — such as scene motion intensity, camera movement, lighting changes, and compression artifacts — can substantially affect the baseline distribution of consecutive frame distances. A fast-moving action sequence naturally exhibits higher cosine distances between frames compared to a static talking-head video, potentially triggering false positives even in authentic regions. Conversely, subtle manipulations in low-motion scenes may produce distance variations below the fixed threshold, leading to false negatives.

This observation suggests that an optimal, video-adaptive threshold selection mechanism is necessary. Rather than applying globally fixed parameters, a more robust approach would dynamically determine thresholds based on each video's intrinsic properties. For instance, the baseline motion statistics of a video could be analyzed to compute a content-aware threshold, or an unsupervised method could be developed to identify statistically significant deviations relative to the video's own temporal dynamics. Future work will explore adaptive thresholding algorithms, potentially leveraging the distribution of Z-scores or distance values within each video to automatically determine the most sensitive and specific cutoff points.

% ============================================================
\section{Conclusion}
% ============================================================

In this paper, we investigated the problem of detecting temporally localized manipulated segments inserted into authentic video streams. We provided a comparative review of existing video forgery datasets and identified the absence of benchmarks that explicitly model this realistic tampering scenario. To address this gap, we introduced a custom-curated, partially manipulated video test set and evaluated two complementary detection frameworks. Our results demonstrate that while supervised frame-by-frame classification (Linear Probe) struggles with low sensitivity, failing to detect the majority of manipulated frames due to cross-domain generalization challenges in authentic contexts, our proposed unsupervised temporal anomaly detection—utilizing cosine distance on DINOv3 features alongside a minimum sensitivity threshold—delivers superior performance. It successfully localizes manipulations while maintaining exceptional robustness, as evidenced by a 95\% video-level accuracy and a near-zero false alarm rate on a 100-video authentic control group. These findings motivate future work on video-specific adaptive thresholding and the systematic construction of larger datasets tailored to this scenario.

\bibliographystyle{IEEEtran}
\bibliography{references}

\end{document}